\documentclass[11pt]{article}

\usepackage[preprint]{acl}

\usepackage{times}
\usepackage{latexsym}

\usepackage[T1]{fontenc}

\usepackage[utf8]{inputenc}

\usepackage{microtype}

\usepackage{inconsolata}

\usepackage{graphicx}

\usepackage{booktabs}
\usepackage{amsmath}
\usepackage{amssymb}
\usepackage{forest}
\usepackage{xcolor}
\definecolor{vc}{HTML}{0072B2}
\definecolor{cc}{HTML}{E69F00}
\usepackage{todonotes}

\definecolor{lilac}{HTML}{B57EDC}

\title{Left-Branching Transformers Excel at Right-Branching Languages: 
Data Shapes Word Order Preferences in Language Models}

\author{
  \textbf{Varvara Arzt\textsuperscript{1}},
  \textbf{Allan Hanbury\textsuperscript{1}},
  \textbf{Terra Blevins\textsuperscript{2}}
\\[0.5em]
  \textsuperscript{1}Faculty of Informatics, TU Wien \\
  \textsuperscript{2}Khoury College of Computer Sciences, Northeastern University
\\[0.3em]
  \small{
    \textbf{Correspondence:} \href{mailto:varvara.arzt@tuwien.ac.at}{varvara.arzt@tuwien.ac.at}
  }
}

\begin{document}
\maketitle
\begin{abstract}
We systematically compare word order preferences in decoder-only language models across 192 artificial languages and typologically diverse natural languages. On artificial languages, models exhibit a left-branching preference that aligns with neither natural language universals nor human word order learning biases. On natural languages, monolingual models show no clear base word order bias at small scales, but as data grows, a preference for right-branching subject-verb-object (SVO) languages emerges while SOV falls behind despite being the most frequent order cross-linguistically. This SVO advantage extends to multilingual models and correlates with language resource level and data quality rather than word order. Thus, the same architecture exhibits opposite preferences on artificial and natural languages, establishing that word order biases observed in practice are data-driven. Since highly-resourced languages are overwhelmingly SVO, these biases risk gradually reducing word order diversity, particularly in languages that productively use multiple word orders, with the widespread adoption of LLMs.\footnote{Code and data are available at \url{https://github.com/kleines-gespenst/word-order-preferences}}

\end{abstract}

\section{Introduction}
\begin{figure}[!t]
  \centering
  \includegraphics[width=\linewidth]{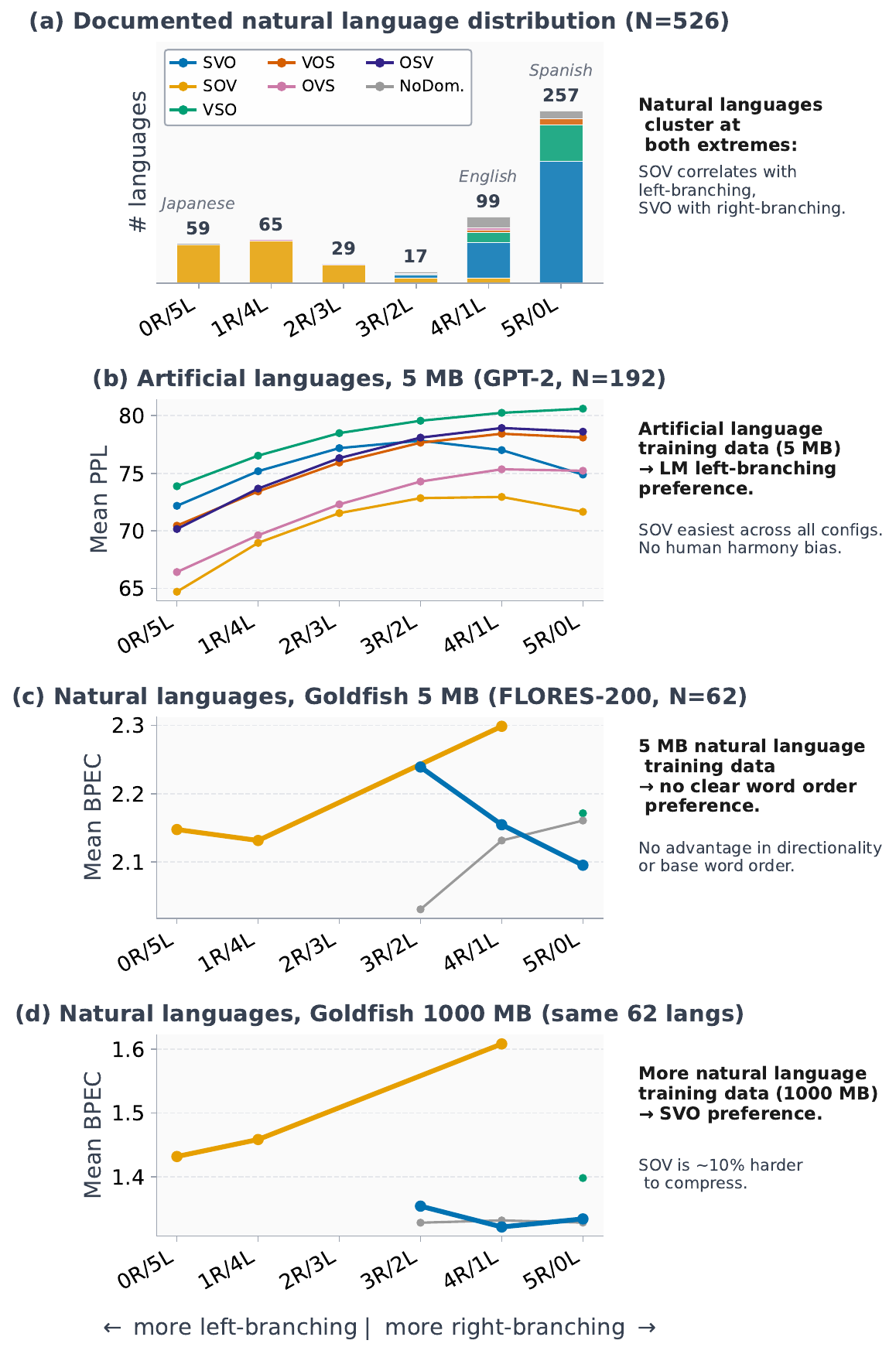}
\caption{Head directionality vs.\ model performance across attested, artificial, and natural languages. $x$-axis: right-branching switch count (\S\ref{sec:meth-artificial-languages}), from fully left-branching (0R/5L) to fully right-branching (5R/0L). $\downarrow$\,=\,better in (b)--(d); colours denote base word order (legend in (a)).
\textbf{(a)}~Attested language frequency by base word order.
\textbf{(b)}~GPT-2 on artificial languages (PPL).
\textbf{(c,\,d)}~Monolingual models on natural languages at two training data
scales (BPEC; §\ref{sec:metrics}).
}
\vspace{-20pt}
  \label{fig:rightness-score}
\end{figure}

Word order is one of the most fundamental dimensions of cross-linguistic variation. Languages vary widely
in how they linearise both clause-level constituents (e.g., SVO in English, SOV in Japanese) and internal constituent orderings, such as the placement of adpositions and adjectives relative to their heads; however, many of these ordering choices tend to correlate \citep{greenberg1963universals,dryer1992greenbergian,dryer_order_2013,verkerk2026enduring} with, for instance, SOV languages typically using postpositions and SVO languages prepositions, placing languages on a scale of \emph{directionality} (Figure~\ref{fig:rightness-score}a) from \emph{left-branching}, where dependents precede their heads (e.g., postpositions), to \emph{right-branching}, where heads come first (e.g., prepositions). Beyond dominant orders, many languages exhibit productive word order variation that serves discourse-pragmatic functions such as topicalisation and focus marking
\citep{hull-dobrovoljc-2025-word}. As language models (LMs) are increasingly deployed across typologically
diverse languages, a natural question arises: do these models develop systematic word order preferences, and
if so, are such preferences architectural or data-driven? This question is especially pressing due
to the potential for LLM-mediated language homogenisation \citep{kew-etal-2024-turning,fitterer-etal-2025-testing} and LLM bias towards more formal writing styles \citep{abdulhai2026llmsdistortwrittenlanguage}, which tend to favour rigid word orders.

Prior work has used artificial languages generated from formal grammars \citep[e.g.,][]{white-cotterell-2021-examining} and synthetic variants of natural languages \citep[e.g.,][]{ravfogel-etal-2019-studying} to study the word order preferences of LMs.
However, artificial languages lack the semantic, morphological, and discourse-level complexity of natural languages, so findings from controlled settings may not generalise; conversely, evaluation on natural languages alone cannot isolate word order from confounds such as morphology, data quality, and script. We combine both approaches at multiple training data scales to address this gap, providing three methodological contributions: \\

\noindent (1) We train models on 192 artificial languages spanning all six base word orders as well as 32 internal constituent orderings, analysing directionality preferences, alignment with cross-linguistic universals, and constituent-level surprisal at multiple training data scales. \\

\noindent (2) We similarly evaluate monolingual and multilingual models trained on typologically diverse natural languages, analysing how word order preferences and constituent surprisal scale with growing data. \\

\noindent (3) By mapping artificial language configurations to natural language typological features and comparing models trained at matched data scales across both paradigms, we unify artificial and natural evaluations of word order preference into a single, controlled comparison, allowing us to disentangle architectural and data-driven biases.\\

We find that on artificial languages, models exhibit a left-branching preference (Figure~\ref{fig:rightness-score}b) that aligns with neither cross-linguistic universals nor human learners' preference for consistent directionality \citep{CULBERTSON201571}. Constituent surprisal analyses suggest this preference may partly reflect the entropy structure of a semantics-free grammar rather than a model-inherent architectural prior. However, these base word order rankings can shift with data volume, and directionality preferences reverse early in training. 

In contrast, while no word order preference is visible at 5\,MB (Figure~\ref{fig:rightness-score}c) for natural languages, we find that SVO languages are easiest and SOV hardest in monolingual models at 1000\,MB (Figure~\ref{fig:rightness-score}d), implicating data quality rather than quantity alone. 
This SVO advantage extends to multilingual models and disappears only when this word order is dominated by very-low-resource languages, as in BLOOM \citep{workshop2023bloom176bparameteropenaccessmultilingual}. 

These findings show that, in practice, models consistently favour the word orders of highly-resourced languages, particularly at scale.
Since highly-resourced languages are overwhelmingly SVO (e.g., English and Chinese; Appendix~\ref{app:resource-word-order}), this creates a precondition for \textit{silent typological homogenisation}: a gradual reduction of word order diversity driven by imbalanced training data. Our results suggest that this precondition is real, robust across model types and training regimes, and in principle addressable through data curation.

\section{Related Work}
\label{sec:related-work}

LM word order preferences have mainly been studied through artificial languages and synthetic variants of natural languages.

\citet{white-cotterell-2021-examining} train LMs on languages generated from a probabilistic context-free grammar (PCFG) varying in branching directionality, finding a preference for left-branching configurations and best performance on both widely attested SOV and rarely attested OVS, suggesting no
alignment with typological frequency; our artificial language experiments build upon their approach
(\S\ref{sec:meth-artificial-languages}). \citet{kuribayashi-etal-2024-emergent} extend this to
cognitively-motivated LMs, finding that models with parsing strategies and memory limitations achieve lower perplexity (PPL) on typologically frequent word orders. \citet{el-naggar-etal-2025-word} adopt
Generalized Categorial Grammar, covering all six S/V/O linearisations, and find that typologically plausible orders
facilitate length generalisation. In emergent communication, \citet{lian-etal-2023-communication} show that neural agents converge from an equal-probability mixed-order language toward a single dominant order
through communicative pressure alone, an effect that strengthens with larger agent group size \citep{lian-etal-2024-nellcom}.

A complementary line of work creates synthetic variants of attested languages. \citet{ravfogel-etal-2019-studying} create synthetic English variants with altered word order and case marking, finding that RNN agreement prediction reflects recency bias. \citet{Bisazza2021FreeOrder} similarly alter word order flexibility
and case marking in English to study neural machine translation, finding that translating flexible-order languages is harder in low-resource settings. \citet{kallini-etal-2024-mission} show that GPT-2 assigns higher
PPL to English variants with disrupted hierarchical structure, though \citet{ziv-etal-2026-biasless} find this does not generalise across nine diverse languages. \citet{clark-etal-2023-cross} find that real word orders distribute information more uniformly than counterfactual alternatives, with the effect strongest for SVO. In concurrent work, \citet{xu2026can} find that disharmonic (i.e., typologically inconsistent in directionality) variants of English and Japanese based on five Greenbergian universals\footnote{Cross-linguistic generalisations about word order co-occurrences
\citep{greenberg1963universals}, e.g., SOV $\leftrightarrow$ postpositions.} are learned more slowly but reach comparable final performance; we test a partly overlapping set of universals (\S\ref{sec:metrics}). 

No prior work, to our knowledge, combines controlled artificial and natural language experiments to analyse word order preferences; we do so here.

\section{Methodology}
\label{methodology}

To disentangle potential architectural biases from effects of training data, we perform parallel experiments on artificial and natural languages, using statistics on attested\footnote{Documented natural languages} languages as a reference for whether model preferences correlate with human biases. Artificial languages allow us to analyse all possible word order permutations across syntactic constructions in isolation, avoiding confounds from typological dimensions like morphological agreement \citep{white-cotterell-2021-examining}. Experiments with natural languages address whether findings on artificial languages generalise, and help establish whether word order preferences can be reliably studied with artificial languages \citep{el-naggar-etal-2025-gcg,ziv-etal-2026-biasless}.
\subsection{Artificial Languages}
\label{sec:meth-artificial-languages}

\begin{figure}[t]
\centering
\small
\renewcommand{\arraystretch}{1.2}
Each language is encoded as \texttt{[Base]-[5-char]}, where the
base specifies the S/V/O order and each character selects a binary
parameter: \texttt{L} = left-branching, \texttt{R} = right-branching.
\vspace{0.2cm}
\begin{tabular}{@{}cll@{}}
\toprule
\textbf{Switch} & \textbf{L (left-branching)} & \textbf{R (right-branching)} \\
\midrule
1 VP\_Comp & $S_{Comp}\ Verb_{Comp}$ & $Verb_{Comp}\ S_{Comp}$ \\
2 Comp     & $S\ Comp$               & $Comp\ S$ \\
3 PP       & $NP\ Postp$             & $Prep\ NP$ \\
4 NP       & $Adj\ Noun$             & $Noun\ Adj$ \\
5 Rel      & $VP\ Rel\ Noun$         & $Noun\ Rel\ VP$ \\
\bottomrule
\end{tabular}
\vspace{0.2cm}
\textbf{Example:} \texttt{SOV-LLLLL} $\approx$ \textbf{Japanese}
\vspace{0.1cm}
\begin{tabular}{@{}ll@{}}
\toprule
\textbf{Feature} & \textbf{Value} \\
\midrule
Base (SOV)   & Subject--Object--Verb \\
Switch \ 1 = L  & complement clause before verb \\
Switch \ 2 = L  & clause before complementiser \\
Switch \ 3 = L  & postpositions \\
Switch \ 4 = L  & adjective before noun \\
Switch \ 5 = L  & prenominal relative clause \\
\bottomrule
\end{tabular}
\vspace{0.2cm}

6 base orders $\times$ $2^5$ configurations = \textbf{192 languages}.
\caption{Language encoding scheme}
\label{fig:switchable-rules}
\end{figure}

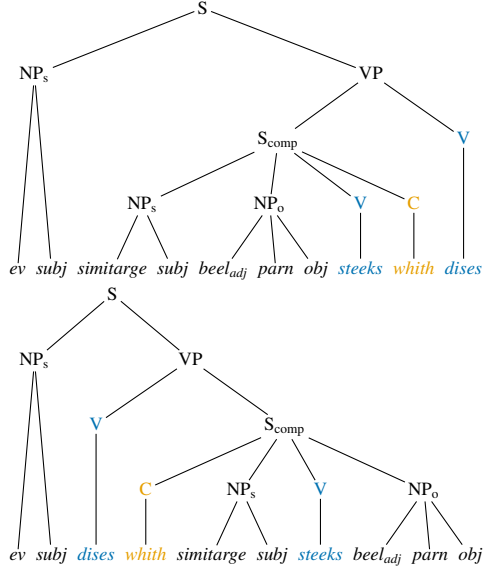
\begin{figure}[t]
\centering\small
\begin{forest}
  for tree={s sep=1.5pt,inner sep=1pt,l sep=6pt,font=\scriptsize},
  where n children=0{font=\scriptsize\itshape,tier=w}{}
  [S
    [NP$_\text{s}$ [ev][subj]]
    [VP
      [S$_\text{comp}$
        [NP$_\text{s}$ [simitarge][subj]]
        [NP$_\text{o}$ [beel$_\text{adj}$][parn][obj]]
        [{\color{vc}V} [{\color{vc}steeks}]]
        [{\color{cc}C} [{\color{cc}whith}]]
      ]
      [{\color{vc}V} [{\color{vc}dises}]]
    ]
  ]
\end{forest}
\hfill
\begin{forest}
  for tree={s sep=1.5pt,inner sep=1pt,l sep=6pt,font=\scriptsize},
  where n children=0{font=\scriptsize\itshape,tier=w}{}
  [S
    [NP$_\text{s}$ [ev][subj]]
    [VP
      [{\color{vc}V} [{\color{vc}dises}]]
      [S$_\text{comp}$
        [{\color{cc}C} [{\color{cc}whith}]]
        [NP$_\text{s}$ [simitarge][subj]]
        [{\color{vc}V} [{\color{vc}steeks}]]
        [NP$_\text{o}$ [beel$_\text{adj}$][parn][obj]]
      ]
    ]
  ]
\end{forest}
 
\vspace{-4pt}
\caption{Same sentence as \texttt{SOV-LLLLL} ($\approx$\,Japanese, top) 
and \texttt{SVO-RRRLR} ($\approx$\,English, bottom): same hierarchical structure, but word order differs. {\color{vc}Verbs} \& 
{\color{cc}complementiser} highlighted; \textit{subj}/\textit{obj} = overt case markers.}
\label{fig:example-trees}
\end{figure}

We generate artificial languages building upon \citet{white-cotterell-2021-examining}. Each language is defined by a base S/V/O order and five binary head-direction switches controlling the ordering of complement clauses, complementisers, adpositions, adjectives, and relative clauses (Figure~\ref{fig:switchable-rules}). Each switch selects between an \textbf{L} (left-branching, head-final) variant, in which the dependent precedes the head, and an \textbf{R} (right-branching, head-initial) variant, in which the head precedes the dependent. A base PCFG with a Zipf-distributed lexicon generates bracketed parse trees; main clause constituents are then deterministically reordered and head-direction switches applied, so that all 192 variants share identical derivation probabilities and differ only in word order. Figure~\ref{fig:example-trees} illustrates this with the same sentence under \texttt{SOV-LLLLL} ($\approx$\,Japanese) and \texttt{SVO-RRRLR} ($\approx$\,English): the hierarchical structure is preserved while the surface linearisation changes.

We extend \citet{white-cotterell-2021-examining} in three ways: (1)~we add an explicit switch for clausal complement order, which \citet{white-cotterell-2021-examining} tie to nominal object order, making them inseparable by design;\footnote{Nominal and clausal object placement are
typologically correlated \citep{dryer1992greenbergian}
but not always identical; e.g., the Agob-Ende-Kawam language has preverbal nominal but postverbal clausal objects \citep[GB135]{grambank2023data}.} (2)~we add a six-way base word order parameter covering all six S/V/O linearisations, as does \citet{el-naggar-etal-2025-word} but via a different formalism. Combining six base orders with $2^5$ switch configurations yields \textbf{192 artificial languages} with fixed word order; and (3)~we scale the vocabulary from ${\sim}$1,400 to 50k pseudowords generated by Wuggy \citep{keuleers-brysbaert-2010-wuggy} from the
most frequent English words in wordfreq \citep{speer-2022-wordfreq}, with Zipfian sampling weights ($\alpha{=}1.0$) per part-of-speech class (details in Appendix~\ref{app:artificial-languages}). 

\subsection{Natural Languages}

\paragraph{Word Order Clustering}
\label{sec:wo-labeling}
To enable comparison between artificial and natural language experiments, we classify each natural language using the same encoding scheme applied to our artificial languages. We obtain word order labels from the typological databases WALS \citep{dryer_order_2013}, Grambank \citep{skirgardGrambankRevealsImportance2023}, and APiCS \citep{apics2013} and map them to our artificial language labels (e.g., English $\rightarrow$ \texttt{SVO-RRRLR}, Japanese $\rightarrow$ \texttt{SOV-LLLLL}). We supplement database labels with manual verification grounded in peer-reviewed publications devoted to individual languages, and use \texttt{lang2vec} \citep{littell-etal-2017-uriel} and URIEL+ \citep{khan-etal-2025-uriel} for language clustering and feature extraction. Languages with non-binary or missing values for a given feature were excluded from our analysis. Full
feature mapping procedure and coverage statistics are given in Appendix~\ref{app:nat-lang-labels}.

We note that typological database classifications are typically based on grammar descriptions rather than corpus frequencies \citep{levshina_word_order_gradient, hull-dobrovoljc-2025-word}, and a single categorical label may obscure gradient word order variation \citep{levshina_word_order_gradient}; see \S~Limitations for further details.

\paragraph{Data}
\label{sec:data}

We evaluate on two datasets. Our primary evaluation set is FLORES-200 \citep{nllbteam2022languageleftbehindscaling}, which provides parallel sentences across a typologically broad set of languages. We also evaluate on the Parallel Universal Dependencies (PUD) treebanks \citep{zeman-etal-2017-conll}, whose languages are all covered by FLORES-200, which both verify our FLORES-200 findings and provide gold-standard syntactic annotations for the per-token surprisal analysis, complementing automatic Stanza \citep{qi-etal-2020-stanza} parses on FLORES-200. Full list of languages used in \S\ref{sec:results} is given in Appendix~\ref{app:eval-languages}.

\subsection{Models}
\label{sec:models}
For artificial languages, we train a separate GPT-2-style decoder-only model on each of the 192 languages to evaluate whether word order preferences
arise from the transformer architecture itself. For natural languages, we primarily focus on monolingual Goldfish models \citep{chang2026goldfishmonolinguallanguagemodels}, GPT-2-based models with a 50k-token vocabulary, scaled in parameter count with training data volume, trained on one of up
to 350 languages in a maximally comparable setup at four training data volumes (5\,MB, 10\,MB, 100\,MB, and 1000\,MB of content-equivalent text per 
language after byte premium scaling; \citealp{chang2026goldfishmonolinguallanguagemodels})\footnote{Byte premium scaling adjusts raw data sizes by UTF-8 byte ratio relative to English \citep{arnett2024bit}: e.g., the 5\,MB Burmese model was trained on 25\,MB raw text (\url{https://huggingface.co/goldfish-models/mya_mymr_5mb}).}, enabling systematic analysis of how word order preferences develop with growing amounts of data. To verify whether these patterns generalise to multilingual training, we additionally evaluate mGPT (1.3B and 13B parameters, ${\sim}$400B tokens across 61 languages with no exact per-language statistics available; \citep{shliazhko-etal-2024-mgpt}), BLOOM (560M, 3B, and 7.1B parameters, 341B tokens across 46 languages, median 527\,MB per language; 
\citep{workshop2023bloom176bparameteropenaccessmultilingual,laurencon2023bigscience}) and XGLM (564M and 1.7B parameters, ${\sim}$500B tokens across 30 languages with exponential upsampling of low-resource languages; \citep{lin2022fewshotlearningmultilinguallanguage}). All three
multilingual models were designed to broaden language coverage beyond English-centric corpora and document their training data composition, allowing us to verify which languages each model was
trained on and to assess potential contamination with FLORES-200 and PUD.

\subsection{Evaluation}
\label{sec:metrics}

We evaluate word order preferences at two levels: clause-level S/V/O linearisation and internal constituent order (the five switches in Figure~\ref{fig:switchable-rules}), aggregating the latter into a directionality scale.

For artificial languages, we report test \textbf{perplexity} (PPL). Following \citet{kallini-etal-2024-mission} and \citet{ziv-etal-2026-biasless}, we additionally analyse training dynamics by tracking test perplexity every 50 steps and computing the area under the PPL curve (AUC); this is not possible for Goldfish models as no intermediate checkpoints are available. We also test whether model preferences align with a subset of \textbf{Greenbergian universals} \citep{greenberg1963universals} testable in our setup: subject-object ordering, base word order and adposition type, adjective placement, and verb-object order with relative clause placement \citep{dryer1992greenbergian} (details in Appendix~\ref{app:universals}). 

For natural languages, where different tokenisers make raw PPL incomparable across languages, we use \textbf{bits-per-English-character} (BPEC; Appendix~\ref{app:bpec}) as defined in \citet{cotterell-bpec}. Throughout, a language is \emph{easier} (\emph{harder}) for a model if it achieves lower (higher) PPL/BPEC. For both artificial and natural languages, we evaluate per-token \textbf{surprisal} \citep{wilcox-etal-2023-testing} to analyse whether tokens with different syntactic functions are differentially predictable across word orders.

\section{Experimental Setup}
\label{sec:experimental-setup}

\subsection{Artificial Languages}
\label{sec:setup-artificial}

We train GPT-2-style decoder-only models using the
HuggingFace Transformers library\footnote{\url{https://huggingface.co/docs/transformers}} (\texttt{GPT2LMHeadModel}) at three training data volumes: the original \citet{white-cotterell-2021-examining} setup (${\sim}$0.5\,MB, 10K sentences), 5\,MB, and 10\,MB. As 0.5\,MB results are comparable to 5\,MB, and we aim for comparability with the monolingual Goldfish models, we focus on the 5\,MB and
10\,MB settings.

The model architecture follows the Goldfish configuration at comparable sizes: 4~layers and 8~attention heads \citep{chang2026goldfishmonolinguallanguagemodels}. Unlike \citet{white-cotterell-2021-examining}, who used a whitespace tokeniser (under which the vocabulary is effectively arbitrary), we train a BPE word-boundary SentencePiece tokeniser \citep{kudo2018sentencepiecesimplelanguageindependent}
with a vocabulary size of 15k (details in Appendix~\ref{app:tokeniser}). Following \citet{white-cotterell-2021-examining}, we train each model on 10 fully disjoint train/dev/test splits generated from independent data and report averaged test perplexity to ensure robustness of results. Implementation details, hyperparameters and compute cost reported in Appendix~\ref{app:model-hyperparams}.

\subsection{Natural Languages}
\label{sec:setup-natural}

We evaluate Goldfish models, available on HuggingFace,\footnote{\url{https://huggingface.co/goldfish-models}} at all four training data sizes and the multilingual models introduced in \S\ref{sec:models} on FLORES-200 \citep{nllbteam2022languageleftbehindscaling} (Appendix~\ref{app:goldfish-coverage}) and PUD \citep{zeman-etal-2017-conll}. Syntactic annotations for the surprisal analysis come from Stanza
\citep{qi-etal-2020-stanza} on FLORES-200 and gold UD annotations on PUD. For multilingual models, we restrict evaluation to the subset of each model's training languages overlapping with FLORES-200 (language counts per model in Table~\ref{tab:multilingual_bpec}).

\section{Results}
\label{sec:results}
We compare LM word order preferences on artificial languages (\S\ref{sec:artificial_results}), where models show a strong left-branching preference misaligned with cross-linguistic universals, and natural languages (\S\ref{sec:natural_results}), where training data composition increasingly shapes model preferences.

\begin{figure*}[t]
  \centering
  \includegraphics[width=\textwidth]{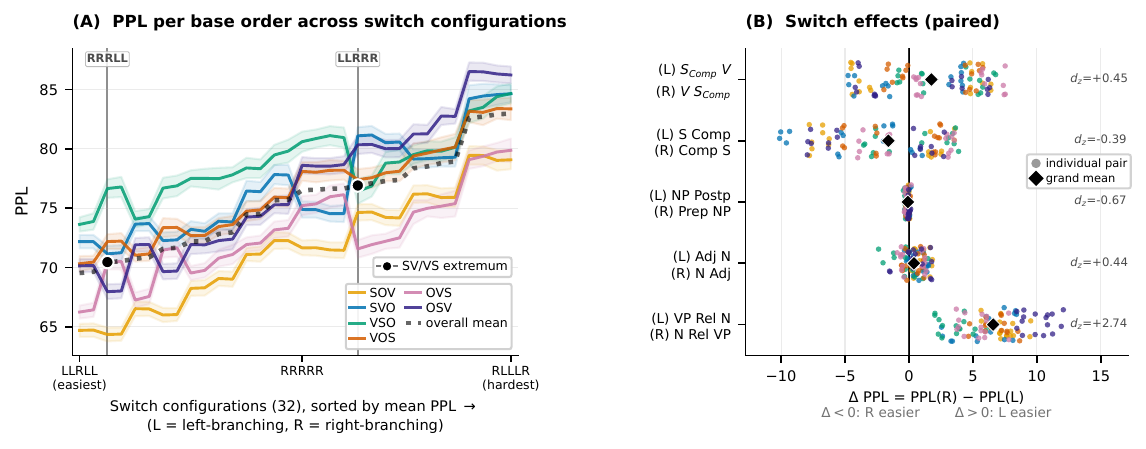}
\caption{
LM preferences on 192 artificial languages (6 base orders $\times$ 32 branching configurations), averaged over 10 runs on 5\,MB data.
\textbf{(A)}~PPL per base order across configurations, sorted by across-order mean (dotted line). Shaded bands: $\pm1$ SD. Vertical markers (\texttt{RRRLL}, \texttt{LLRRR}): SV and VS base orders respond inversely to clause-level switches.
\textbf{(B)}~Paired switch effects ($\Delta\mathrm{PPL}$) for configuration pairs differing only in the target switch. $\blacklozenge$: overall mean, where $>0=$ left-branching is preferred. Switch effect sizes $d_z$ are all significant ($p<.001$).
}
  \label{fig:fig-artificial-5mb}
\end{figure*}

\subsection{Artificial Languages}
\label{sec:artificial_results}

On artificial languages, models consistently prefer left-branching configurations, conflicting with cross-linguistic universals, and base word order rankings shift unstably with data volume.

\paragraph{Directionality Preferences}
Figure~\ref{fig:rightness-score}b shows that the mean PPL across settings increases consistently with the number of right-branching switches in an artificial language across all six base word orders, with only slight decreases in PPL for majority right-branching settings in SVO and SOV languages. This preference for left-branching configurations corroborates prior work \citep{white-cotterell-2021-examining,kuribayashi-etal-2024-emergent,el-naggar-etal-2025-gcg}. 

However, the top-ranked configuration for each base order has no close natural language counterpart,\footnote{A partial exception is SVO-RRRLL ($\approx$\,Mandarin Chinese), though the prenominal relative clause (switch~5\,=\,L) is very rare in VO languages \citep{dryer1992greenbergian, comrie2008prenominal}.} suggesting that transformers do not mirror human word order learning biases. In contrast, attested languages cluster at both extremes of the directionality scale (Figure~\ref{fig:rightness-score}a), consistent with the human harmony bias toward consistent directionality \citep{CULBERTSON201571}.

The magnitude of the left-branching preference also varies substantially across individual switches (Figure~\ref{fig:fig-artificial-5mb}B). Switch~5 (relative clause) dominates: all sixteen left-branching relative clause configurations are easier than the right-branching ones, at $\Delta$PPL of 4--7 points on 5\,MB. This strong preference for prenominal relatives runs counter to the typological distribution, where postnominal relatives outnumber prenominal ones approximately 4:1 \citep{dryer1992greenbergian,dryer_order_2013}. Other switches show smaller but consistent trends; for example, Switch~4 (pre- or postnominal adjectives) shows a moderate preference for adjectives preceding the noun. 
All five paired $t$-tests on the effect of internal constituent switches are significant even after a strict Holm--Bonferroni correction to remove false positives ($p<.001$). More broadly, switches that reorder clause-level constituents (relative clauses, complement clauses) produce larger perplexity differences than switches that rearrange phrase-internal elements (adjective--noun, adposition--NP), suggesting greater sensitivity to the direction of long-range dependencies.

\paragraph{Base Word Order} Model preferences of base word orders are not stable across training data volumes and also do not reflect typological frequency (Appendix Figure~\ref{fig:combined-encoding-basewo}). At 5\,MB, SOV ranks first and VSO last in mean PPL across configurations (Figure~\ref{fig:fig-artificial-5mb}A), consistent with \citet{white-cotterell-2021-examining}. At 10\,MB, the three VS orders rank 1--3 and the three SV orders rank 4--6, with SVO the hardest (Appendix Figure~\ref{fig:fig-artificial-10mb}), which is opposite the trend we see on natural languages (where SVO is easiest to model at scale; \S\ref{sec:natural_results}).

Internal constituents also interact with base word order, with individual base order curves crossing over for specific switch settings (Figure~\ref{fig:fig-artificial-5mb}A): SV bases (SVO, SOV, OSV) and VS bases (VSO, VOS, OVS) respond differently to individual switches. Right-branching verb complements and complementisers (switches~1--2\,=\,R) selectively benefit SV bases by up to 5.5~PPL points (vertical marker \texttt{RRRLL} in Figure~\ref{fig:fig-artificial-5mb}A), while right-branching relative clauses (switch~5\,=\,R) reverse this, favouring VS bases by up to 3.8~points (vertical marker \texttt{LLRRR}). This observed SV/VS split does not align with the VO/OV distinction underlying most implicational word order universals \citep{greenberg1963universals,dryer1992greenbergian,dryer_order_2013}, and direct tests of typologically expected combinations confirm this: the tested universals are either not reflected (adposition, adjective placement) or significant in the opposite direction (subject--object order and relative clause placement, at 10\,MB). These results indicate that the inductive biases demonstrated by transformers on artificial languages do not match cross-linguistic word order universals.

\paragraph{Training Dynamics}

\begin{figure}[t]
  \centering
  \includegraphics[width=\linewidth]{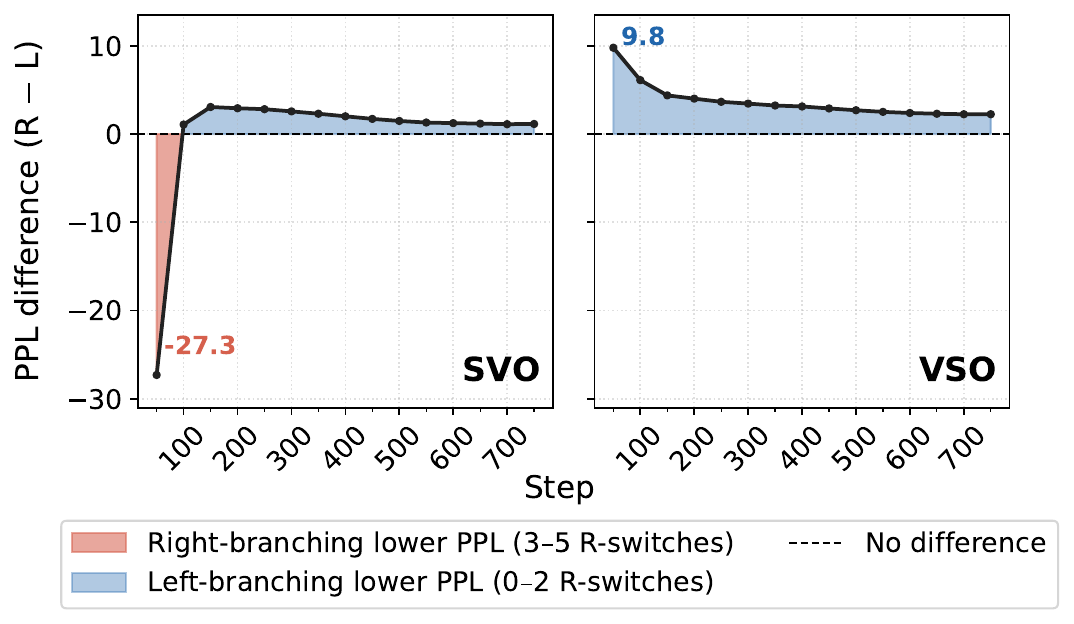}
  \caption{PPL $\Delta$ (right $-$ left branching) during training on 5\,MB artificial data.
  \textbf{SVO}: right-branching is initially easier before reversing to a left-branching advantage by step~100.
  \textbf{VSO}: left-branching always dominates.}
  \label{fig:ppl_diff_2panel}
\end{figure}

The model's directionality preference shifts in the earliest phase of training, again following the SV/VS split (Figure~\ref{fig:ppl_diff_2panel} for SVO and VSO, with other orders following this trend in Appendix Figure~\ref{fig:ppl_diff_5mb}). 
For SV base orders, right-branching configurations are initially easier (by up to 28~PPL points) before fully reversing to a left-branching advantage by step~100 (epoch~${\sim}$2.6); for VS orders, left-branching dominates from the first checkpoint. Throughout training, the internal constituent order switches rather than the base S/V/O order drive the PPL spread, with convergence speed and final PPL strongly correlated ($\rho > 0.89$; Appendix Figure~\ref{fig:convergence-ranking}).
Moreover, the early reversal for SV languages echoes the preference shift observed by \citet{el-naggar-etal-2025-gcg} and indicates that directionality preferences may be sensitive to training dynamics and data quantity, which should be considered when comparing results across studies.

\subsection{Natural Languages}
\label{sec:natural_results}

While LMs show no word order preference in small training data scales on natural languages, an SVO advantage emerges at scale in both mono- and multilingual models, correlating with language resource level and data quality rather than word order type.

\begin{table}[t]
\centering
\small
\setlength{\tabcolsep}{4pt}
\begin{tabular}{@{}lrrrr@{}}
\toprule
\textbf{Base order} & \textbf{5\,MB} & \textbf{10\,MB} & \textbf{100\,MB} & \textbf{1\,GB} \\
\midrule
SVO\,{\scriptsize(32)}    & \textbf{2.13}\,{\scriptsize(.15)} & \textbf{2.00}\,{\scriptsize(.15)} & \textbf{1.55}\,{\scriptsize(.08)} & \textbf{1.32}\,{\scriptsize(.06)} \\
SOV\,{\scriptsize(22)}    & \textbf{2.13}\,{\scriptsize(.10)} & 2.02\,{\scriptsize(.10)} & 1.64\,{\scriptsize(.11)} & 1.45\,{\scriptsize(.12)} \\
VSO\,{\scriptsize(3)}     & 2.20\,{\scriptsize(.07)} & 2.06\,{\scriptsize(.09)} & 1.64\,{\scriptsize(.06)} & 1.42\,{\scriptsize(.06)} \\
NoDom\,{\scriptsize(11)}  & \textbf{2.13}\,{\scriptsize(.11)} & \textbf{2.00}\,{\scriptsize(.10)} & 1.56\,{\scriptsize(.08)} & 1.33\,{\scriptsize(.05)} \\
\midrule
\textit{SVO--SOV} $\delta$ & $0.00$ & $-0.10$ & $-0.60^{***}$ & $-0.76^{***}$ \\
\bottomrule
\multicolumn{5}{@{}l@{}}{\scriptsize At 5\,MB: SVO\,=\,2.1288,
SOV\,=\,2.1289, NoDom\,=\,2.1284.} \\
\end{tabular}
\caption{Median BPEC and (interquartile range; IQR) for base word orders of 68 monolingual Goldfish models (evaluated on FLORES-200, lang.\
count in parentheses). $\downarrow$\,=\,better, \textbf{bold}\,=\,best per column. Bottom row: SVO--SOV effect size (Cliff's $\delta$); order preference is absent at 5\,MB (Mann--Whitney $U{=}353$, $p{=}.99$) and large by 1\,GB
($U{=}84$, $p{<}10^{-5}$). ${}^{***}p{<}.001$.}
\label{tab:median-bpec-by-size}
\end{table}

\paragraph{Directionality Preferences}

Figure~\ref{fig:rightness-score}c demonstrates that at 5\,MB, LMs trained on natural languages exhibit no clear directionality preference; however, by 1000\,MB, right-branching SVO languages become easiest (Figure~\ref{fig:rightness-score}d; intermediate scales in Appendix Figure~\ref{fig:rightness-goldfish}).\footnote{Here we restrict to the 68 languages with Goldfish models at all four training data sizes. When all available languages at each size are included (Appendix Figure~\ref{fig:goldfish_all_langs_directionality}), SVO even underperforms SOV at small scales.}
This may reflect the same left-branching inductive bias observed in artificial languages, overridden by data effects in natural languages as training data grows.

\paragraph{Base Word Order Preferences across Scale}

This shift across data scales is also visible at the level of base word order. At 5\,MB, monolingual Goldfish models show no preference: SVO, SOV, and NoDominant all centre around BPEC 2.13, with no significant difference between
orders (Table~\ref{tab:median-bpec-by-size}). As data grows,
an SVO advantage emerges and strengthens monotonically (SVO--SOV Cliff's
$\delta{=}{-}0.76$ at 1\,GB):
by 1\,GB, SOV incurs ${\sim}$10\% higher compression cost
($\Delta$BPEC\,$\approx$\,0.13), despite being the most frequent order
cross-linguistically (Appendix Figure~\ref{fig:combined-encoding-basewo}). All four model sizes are evaluated on the same 68 languages, so the shift is driven solely by training data volume, not language selection. The same pattern holds on PUD (details in Appendix Table~\ref{tab:pud_bpec_by_order}).

Because word order correlates with other language properties, the observed SVO advantage may also be an artefact of a confound such as morphological complexity \citep{arnett-bergen-2025-language} rather than a genuine word order preference. To test this, we fit a linear mixed-effects model on the BPEC of the SVO/SOV languages in Figure~\ref{fig:rightness-score}c,d, with a word order $\times$ size interaction, per-language random intercepts, and three covariates: morphological complexity (subword MATTR; \citealt{tatariya-etal-2025-interplay}), language resourcedness \citep{joshi-etal-2020-state}, and training data composition (OSCAR web-crawl vs.\ curated share; Appendix~\ref{app:mixed-effects}). With all three covariates included, the growth of the SVO--SOV gap with data survives every control ($\beta$: ${+}0.049 \to {+}0.038$, $p{\le}.001$; Appendix Table~\ref{tab:mixed-effects}). Language resourcedness is the only covariate with an observable effect on this growth (${\sim}$20\%), but we note it is often strongly correlated with other, likely important factors we did not control for, such as data quality, which disproportionately hurts low-resource languages \citep{oladipo-etal-2023-better,wang-etal-2025-multilingual-language}\footnote{Within the SVO group of Goldfish models, the easiest language shifts with scale: Indonesian has the lowest BPEC at 5 and 10\,MB, with English just behind, but English leads at larger scales, consistent with higher-resource languages benefiting more from cleaner data as data grows.} (see \S~Limitations). Script is a further possible confound, since SOV and SVO languages differ systematically in writing system: in FLORES-200, SOV languages are non-alphabetic far more often than SVO (71\% vs.\ 19\%). Restricting to segmental alphabets, however, the gap still grows with data (29 SVO / 7 SOV: near zero at 5\,MB to $+0.057$ at 1\,GB, 95\% CI $[+0.011, +0.131]$). Beyond the aggregate BPEC results, per-constituent surprisal further confirms that preferences are data-driven: the ranking reverses from S~$<$~O~$<$~V in semantic-free artificial to V~$<$~O~$<$~S in natural languages both on FLORES-200 and PUD data, with the same architecture yielding opposite profiles depending on whether semantic dependencies are present. 

\paragraph{Multilingual Models}

\begin{table}[t]
\centering
\footnotesize
\setlength{\tabcolsep}{4pt}
\begin{tabular}{@{}lcccc@{}}
\toprule
Model & SVO & SOV & NoDom & $\delta$ \\
\midrule
BLOOM-560m & 3.62\,{\scriptsize(2.86)} & \textbf{2.59}\,{\scriptsize(.59)} & — & $+.26$ \\
BLOOM-3b   & 2.99\,{\scriptsize(2.20)} & \textbf{1.88}\,{\scriptsize(.34)} & — & $+.27$ \\
BLOOM-7b1  & 2.81\,{\scriptsize(2.06)} & \textbf{1.82}\,{\scriptsize(.29)} & — & $+.25$ \\
\midrule
XGLM-564M & \textbf{1.46}\,{\scriptsize(1.06)} & 1.89\,{\scriptsize(4.69)} & 1.47\,{\scriptsize(.63)} & $-.49^{*}$ \\
XGLM-1.7B & \textbf{1.41}\,{\scriptsize(.88)}  & 1.78\,{\scriptsize(5.12)} & 1.42\,{\scriptsize(.54)} & $-.53^{*}$ \\
\midrule
mGPT-1.3B & \textbf{1.53}\,{\scriptsize(.28)} & 1.74\,{\scriptsize(.25)} & 1.64\,{\scriptsize(.31)} & $-.43^{*}$ \\
mGPT-13B  & \textbf{1.44}\,{\scriptsize(.25)} & 1.68\,{\scriptsize(.18)} & 1.54\,{\scriptsize(.24)} & $-.44^{*}$ \\
\bottomrule
\end{tabular}
\caption{Median FLORES-200 BPEC and (IQR) by base word order for multilingual models. Language counts (SVO/SOV/NoDom): BLOOM 28/13/0, XGLM 14/10/3, mGPT 25/22/9. $\downarrow$\,=\,better, \textbf{bold}\,=\,best per model. Last column: SVO--SOV effect size (Cliff's $\delta$); SVO is preferred by XGLM and mGPT ($\delta{<}0$) but not BLOOM ($\delta{>}0$), whose SVO group is very-low-resource and heterogeneous (wide SVO IQR). ${}^{*}p{<}.05$.}
\label{tab:multilingual_bpec}
\end{table}

The SVO advantage and the right-branching directionality
preference both extend to XGLM and mGPT (Table~\ref{tab:multilingual_bpec}; Appendix Figure~\ref{fig:multilingual_combined}), where SVO groups are dominated by highly-resourced Indo-European languages.\footnote{On an alphabetic script subset the multilingual SVO advantage persists in mGPT (22 SVO / 10 SOV, SVO vs.\ SOV Mann--Whitney $p{=}.027$ at 1.3B, $.030$
at 13B; Cliff's $\delta{\approx}{-}0.50$). BLOOM and XGLM have too few alphabetic SOV languages (1--2) to test.} BLOOM is the exception: 21 of its 28 SVO languages are Niger-Congo languages, most very-low-resource, with a median training data volume of 1.7\,MB, compared to 3.7\,GB for all other languages\footnote{The median for the remaining non-Niger-Congo SVO languages in BLOOM is 79\,GB.}, inflating the SVO group's median BPEC and masking the advantage. More broadly, SVO languages receive substantially more training data than SOV languages across all three model families: in BLOOM, the median SVO/SOV ratio is 11$\times$ \citep{laurencon2023bigscience}, reduced to 5$\times$ in XGLM\footnote{\url{https://huggingface.co/facebook/xglm-564M}} after upsampling. The
preference thus reflects resource scale rather than word order. All three model families (mGPT, XGLM, and BLOOM) were explicitly designed to broaden language coverage beyond English-centric corpora, yet the SVO advantage persists wherever the SVO group is not dominated by very-low-resource languages; in BLOOM, the wide interquartile range (IQR) of SVO languages (Table~\ref{tab:multilingual_bpec}, Appendix Figure~\ref{fig:multilingual_combined}) reflects the heterogeneity between the high-resource non-Niger-Congo and the very-low-resource Niger-Congo languages within this group. 

\subsection{Comparison of Artificial and Natural Language Results}
\label{sec:synthesis}

At small training scales, artificial and natural language results partially align: models show a left-branching advantage and no clear base word order preference in both settings. As data grows, however, the two settings consistently diverge across base word orders, internal constituent switches, and constituent-level surprisal trends. 

The left-branching advantage is stable across data sizes in artificial languages but vanishes in natural languages at scale, where right-branching SVO languages become easiest. 
Base word order rankings also reverse; SVO is hardest in artificial languages at 10\,MB (Appendix Figure~\ref{fig:fig-artificial-10mb}) but easiest in natural languages at 1\,GB (Table~\ref{tab:median-bpec-by-size}). Per-constituent surprisal also flips, consistent with semantic selectional restrictions reshaping predictability in natural languages \citep{futrell-2019-information}. 
These contrasts across artificial and natural languages establish that word order preferences in transformer LMs are shaped by data rather than architecture; our multilingual results further reinforce this, with the models' SVO preference disappearing only when very-low-resource languages dominate this word order. Even monolingual Goldfish models trained on content-equivalent amounts of data per language \citep{arnett2024bit} favour the SVO order most commonly seen in highly-resourced languages at scale (Appendix~\ref{app:resource-word-order}), suggesting that data quantity alone does not fully explain this preference. 
Instead, the SVO advantage arises from properties of the training data beyond word order: it survives controls for morphological complexity, script, and, in the coarse form we can measure, training data composition, with only language resourcedness partly explaining it (\S\ref{sec:natural_results}), which is itself a proxy for data quality \citep{oladipo-etal-2023-better,wang-etal-2025-multilingual-language}; finer aspects of data quality we leave to future work (see \S~Limitations).  

\section{Conclusion}

We present a systematic analysis of word order preferences in decoder LMs across artificial and natural languages at multiple training data scales, combining both to disentangle architectural biases from data-driven preferences. On artificial languages, models prefer left-branching configurations with no alignment with cross-linguistic universals, and SVO languages become hardest at scale; on natural languages, no preference emerges at small scales, but LMs exhibit a preference for SVO languages as data grows while SOV falls behind.
These preferences are data-driven: the same architecture exhibits opposite rankings on artificial and natural languages, and the preference for SVO order in natural languages correlates with resource level and data quality rather than word order type. Crucially, even where architectural biases exist (e.g., the left-branching preference on artificial languages), they are overridden by data effects at scale.

Since highly-resourced languages are overwhelmingly SVO, this data-driven advantage has real typological consequences. Given that 1\,GB of training data suffices for the SVO preference to emerge even in monolingual models, the effects we study are likely more pronounced in frontier models where the training data quantity and quality gaps between high- and low-resource languages are even larger. This creates a precondition for what we term \textit{silent typological homogenisation}: a hypothesised gradual reduction of word order diversity in languages that productively use multiple base word orders towards that of the dominant languages. To confirm this hypothesis and better understand the implications of LM word order preferences, we argue that extending this analysis to larger models and spoken language data, testing whether finetuning or downstream task application alters word order preferences \citep{belinkov-glass-2019-analysis}, and applying mechanistic interpretability to identify how the observed order preferences are encoded \citep{olah2022mechinterp} remain important directions for future work.



\section*{Limitations}
\label{sec:limitations}
The scope of our experiments is necessarily limited. Our PCFG production rules, following
\citet{white-cotterell-2021-examining} and \citet{kuribayashi-etal-2024-emergent}, cover a limited subset of
the typological landscape: relative clause placement, for instance, cannot be fully captured by a binary switch given the range of attested strategies \citep{dryer_order_2013}. Our experiments focus on the GPT-2 architecture, consistent with prior work \citep{white-cotterell-2021-examining,kallini-etal-2024-mission} and motivated by direct comparability with Goldfish models \citep{chang2026goldfishmonolinguallanguagemodels}, which share the same architecture and hyperparameters. We did not evaluate larger multilingual models beyond mGPT, BLOOM, and XGLM due to
insufficient transparency regarding per-language training data composition, which is essential for our analysis of resource-level effects. Our language sample may also be biased toward Indo-European languages, which are overrepresented in both typological databases and training data.

Our evaluation data and metrics also carry certain limitations. We evaluate on written, relatively formal corpora (FLORES-200, PUD), but spoken language shows substantially greater word order flexibility \citep{hull-dobrovoljc-2025-word}, and register and genre strongly influence constituent ordering, so our results may not generalise to spoken or informal registers. FLORES-200 and PUD consist of translated text, which may introduce translationese biases
\citep{koppel-ordan-2011-translationese}. Metrics such as PPL and BPEC, while standard, measure overall compression rather than word order preferences directly; behavioural analysis of model outputs would provide more direct evidence.

Our data quality analysis also relies on the observed correlation between language resourcedness \citep[which we find the SVO advantage tracks]{joshi-etal-2020-state} and data quality. While higher-resource languages tend to have cleaner, more diverse corpora \citep{kreutzer-etal-2022-quality,oladipo-etal-2023-better,wang-etal-2025-multilingual-language}, and web-crawled data is noisiest for low-resource languages \citep{kreutzer-etal-2022-quality}, our training data composition variable (OSCAR web-crawl share) remains a coarse proxy, as we do not have direct annotations for data genre or quality. Isolating which finer aspects of data quality drive the effect requires per-language annotation of register, source, and corpus cleanliness, which we leave to future work.

Finally, our analysis relies on categorical word order labels from databases such as WALS \citep{dryer_order_2013} and Grambank \citep{skirgardGrambankRevealsImportance2023}, which encode features in simplified, often binary ways. WALS feature 81A classifies dominant word order based on main clause order alone, with no representation of variability across clause types or the pragmatic conditions under which ordering alternations arise. This can introduce noise: Modern Standard Arabic is classified as VSO based
on a 1958 source, whereas more recent work documents flexible VSO/SVO alternation \citep{aradgreshler2017seeking,
himmelreich2023feature}; German instead is classified as having no dominant order in WALS despite arguably SVO-dominant main clauses, the very criterion WALS uses for classification, while exhibiting SOV in subordinate clauses and clause-final non-finite verbs in auxiliary and modal constructions. Similarly, Russian is classified as SVO-dominant and Belarusian as having no dominant order in WALS, despite comparable word order flexibility in UD corpora \citep{nivre-etal-2020-universal}. Word order variation is arguably gradient rather than categorical \citep{levshina_word_order_gradient}, and a single label may obscure the flexibility of a language's actual word order distribution; even languages labeled with a single dominant order contain sentences with alternative orderings in the evaluation data. Resources such as APiCS partially address this by encoding word order frequencies, but the underlying data are themselves difficult to obtain: reliable estimates require representative corpora,
consistent annotation standards across languages, and large-scale annotation by linguists with near-native competence, none of which is straightforward to achieve. More broadly, many typological features are not binary: Afrikaans, for example, has prepositions, postpositions, and circumpositions.

\bibliography{custom}


\appendix
\section{Appendix}
\label{sec:appendix}

\subsection{Artificial Language Details}
\label{app:artificial-languages}

\paragraph{Vocabulary}
Source words were drawn from the most frequent English words 
in the \texttt{wordfreq} library \citep{speer-2022-wordfreq} and 
used to generate pseudowords with Wuggy 
\citep{keuleers-brysbaert-2010-wuggy}. The resulting vocabulary 
contains ${\sim}$50k wordforms across five open-class categories: 
nouns inflecting for number (50\%), adjectives (10\%), and three 
verb classes (intransitive, transitive, complement-taking) 
inflecting for number and tense (40\% combined). Closed-class 
items comprise 4 prepositions, 7 pronouns, and single invariant 
tokens for the subordinating complementiser, relativiser, 
coordinating conjunction, and overt subject and object case 
markers. This POS inventory follows 
\citet{white-cotterell-2021-examining} and subsequent work 
\citep{kuribayashi-etal-2024-emergent,el-naggar-etal-2025-word}; 
fully replicating natural language POS complexity is infeasible within a controlled PCFG setup. The noun-heavy 
distribution (50\%) reflects this restricted inventory and aligns with corpus-based POS statistics 
\citep{hudson-1994}. Within each open-class category, items are 
assigned Zipfian sampling weights ($\alpha{=}1.0$) by frequency 
rank; closed-class items are sampled uniformly.

\paragraph{Training Corpus}
Each language variant was used to generate 5~MB and 10~MB of training text 
by sampling sentences from the PCFG, compared to 10,000 sentences 
in \citet{white-cotterell-2021-examining}. The PCFG allows 
recursive expansion, with a maximum of 400 expansions per 
sentence to bound recursion depth, following 
\citet{white-cotterell-2021-examining}. Generated sentences have 
an average length of 12 tokens.

\paragraph{Tokeniser}
\label{app:tokeniser}
We train a shared SentencePiece BPE tokeniser \citep{kudo2018sentencepiecesimplelanguageindependent} (15k vocabulary) on the generated corpus in word-boundary mode, so merges never cross word boundaries. Because all 192 languages are permutations of the same sentences with identical vocabulary and word frequencies, the subword inventory depends only on the shared word list, not on word order: retraining the tokeniser separately on each grammar yields identical 15{,}000-piece vocabularies. We also enable \texttt{add\_dummy\_prefix} so that sentence-initial words receive the same token IDs as mid-sentence occurrences, avoiding position-based bias. To confirm our results are not tokeniser-driven, we additionally retrained all 192 models (single seed) with a unigram tokeniser (as used in the Goldfish models; \citealp{chang2026goldfishmonolinguallanguagemodels}) and obtained comparable results: the difficulty ordering is nearly unchanged (Spearman $\rho{=}0.94$ at 5\,MB, $0.91$ at 10\,MB), all five switch effects keep their direction, and the base-order extremes are preserved.

\paragraph{Model and Hyperparameters}
\label{app:model-hyperparams}
We fully reproduced the results of 
\citet{white-cotterell-2021-examining} before conducting our 
experiments. We then made the following modifications.
We adopt the Goldfish architecture 
\citep{chang2026goldfishmonolinguallanguagemodels} to match the 
model size used in our natural language experiments, enabling 
direct comparison (full hyperparameters in 
Table~\ref{tab:hyperparams}). Our models match the Goldfish configuration in layers, attention heads, and embedding dimensions, differing only in vocabulary size (15k vs 50k); prior work with a much smaller vocabulary (${\sim}$1{,}400 pseudowords; \citealp{white-cotterell-2021-examining,kuribayashi-etal-2024-emergent}) recovers the same trends, so our results are not vocabulary-size-driven. We set weight decay to 0, as the 
default value of 0.01 led to over-regularisation at our model 
size and data volume, roughly doubling perplexity in pilot 
experiments. We reduce warmup steps to 10\% of total training 
steps. We cap training at 20 epochs rather than the 10,000 
update steps used by \citet{white-cotterell-2021-examining}, 
which on their data volume results in over 100 epochs. In pilot 
experiments on 5\,MB and 10\,MB corpora, we observed 
train--validation loss crossover at epochs ${\sim}$10 and ${\sim}$16, respectively, with validation loss plateauing or increasing 
beyond epoch 50. Limiting training to 20 epochs is consistent 
with \citet{muennighoff2023scaling}, who find diminishing gains beyond 16 epochs of data repetition; with 
\citet{chang2026goldfishmonolinguallanguagemodels}, who train 
Goldfish models for 10 epochs; and with 
\citet{kuribayashi-etal-2024-emergent}, who use 10 epochs in a similar setup with a smaller training corpus. We evaluate at step level and select the checkpoint 
with the lowest validation loss. We limit artificial language experiments to 5 and 10\,MB to match the Goldfish architecture at these sizes; at larger scales, Goldfish uses more layers, precluding direct comparison. Artificial languages are also easier to model than natural.

\begin{table}[ht]
\centering
\small
\begin{tabular}{ll}
\toprule
\textbf{Hyperparameter} & \textbf{Value} \\
\midrule
\multicolumn{2}{l}{\textit{Model}} \\
Architecture        & GPT-2 (decoder-only) \\
Layers              & 4 \\
Attention heads     & 8 \\
Hidden size         & 512 \\
FFN inner size      & 2,048 \\
Dropout             & 0.1 \\
Activation          & ReLU \\
Tied embeddings     & Yes \\
Parameters          & 20.6M \\
\midrule
\multicolumn{2}{l}{\textit{Tokeniser}} \\
Type                & SentencePiece BPE (word-boundary) \\
Vocabulary size     & 15,000 \\
\midrule
\multicolumn{2}{l}{\textit{Data}} \\
Train / Dev / Test  & 80 / 10 / 10 \\
\midrule
\multicolumn{2}{l}{\textit{Optimisation}} \\
Framework           & HuggingFace Transformers \\
Optimiser           & Adam ($\beta_1{=}0.9$, $\beta_2{=}0.999$, $\varepsilon{=}$1e-6) \\
Learning rate       & 1e-4 \\
LR schedule         & Linear decay with warmup \\
Warmup              & 10\% of total steps \\
Weight decay        & 0 \\
Gradient clipping   & 1.0 \\
Sequence length     & 512 tokens \\
Batch size          & 64 sequences (4 $\times$ 16 grad.\ accum.) \\
Max epochs          & 20 \\
Precision           & bf16 \\
\bottomrule
\end{tabular}
\caption{Model and training hyperparameters for artificial 
language experiments.}
\label{tab:hyperparams}
\end{table}

\paragraph{Compute Cost}
\label{app:compute}

For artificial languages, we trained 1,920 models (192 grammars 
$\times$ 10 splits) at each of two data sizes (5\,MB and 10\,MB) 
on NVIDIA A100 GPUs. A single model takes approximately 
2.5~minutes (5\,MB) and 5~minutes (10\,MB), totalling 
approximately 252 A100 GPU-hours with a wall-clock time of 
approximately 2 days using parallelised Slurm array jobs.

\subsection{Natural Language Details}
\label{app:natural-languages}

\paragraph{Word Order Labeling}
\label{app:nat-lang-labels}
\label{app:lang-classification}
We map each switch in our artificial grammar 
(Figure~\ref{fig:switchable-rules}) to attested typological 
features: \textbf{Base word order}: WALS~81A; \textbf{Switch~1} 
(complement clause order): Grambank~GB135; \textbf{Switch~2} 
(complementiser position): Grambank~GB421; \textbf{Switch~3} 
(adposition order): WALS~85A; \textbf{Switch~4} (adjective--noun 
order): WALS~87A, Grambank~GB193; \textbf{Switch~5} (relative 
clause--noun order): WALS~90A. Features were drawn from \texttt{lang2vec} \citep{littell-etal-2017-uriel}, which aggregates WALS, SSWL, and Ethnologue. Where \texttt{lang2vec} lacked coverage, we consulted WALS \citep{dryer_order_2013} directly, then Grambank \citep{skirgardGrambankRevealsImportance2023} and APiCS \citep{apics2013} via URIEL+ \citep{khan-etal-2025-uriel}. FLORES-200 covers 200 languages, with script alternatives for four (204 total). Of these, 52 lacked a base word order label in WALS, Grambank, and APiCS; we supplemented these using peer-reviewed articles for specific languages, corroborated where possible by corpus-based labels from UD treebanks \citep{nivre-etal-2020-universal}, bringing coverage to 194 languages. Of the 194, 132 have binary (L/R) values for all five switches. Languages with non-binary or missing values were excluded from the corresponding analysis.

\paragraph{Evaluation Languages}
\label{app:eval-languages}
Tables~\ref{tab:multilingual_languages}--\ref{tab:multilingual_languages_2} list all 103 FLORES-200 languages with a base word order label that appear in at least one evaluation set and form the basis of all analyses in \S\ref{sec:results}: 68 Goldfish languages available at all four training sizes,\footnote{Larger sets are available at individual sizes (e.g., 130 at 5\,MB) but are not used in \S\ref{sec:results} to ensure that the same languages are compared across all four training scales, enabling attribution to data volume rather than language sample changes.} 46 BLOOM, 30 XGLM, and 56 mGPT training languages; 20 of the 103 also have a PUD treebank.

\paragraph{Goldfish Model Coverage on FLORES-200}
\label{app:goldfish-coverage}

We evaluated Goldfish models on 183 of FLORES-200 languages, requiring a strict one-to-one match between FLORES codes and dedicated Goldfish models; we do not adopt the cross-variety substitutions in \citet{chang2026goldfishmonolinguallanguagemodels} (e.g.\ \texttt{awa\_Deva}~$\to$~\texttt{hin\_Deva}), as word order differences between varieties could confound our analysis. The 21 excluded varieties (\texttt{ace\_Arab}, \texttt{acm\_Arab}, \texttt{acq\_Arab}, \texttt{aeb\_Arab}, \texttt{ajp\_Arab}, \texttt{arb\_Latn}, \texttt{ars\_Arab}, \texttt{ary\_Arab}, \texttt{awa\_Deva}, \texttt{bjn\_Arab}, \texttt{kam\_Latn}, \texttt{kas\_Arab},
\texttt{min\_Arab}, \texttt{mni\_Beng}, \texttt{npi\_Deva}, \texttt{nus\_Latn}, \texttt{ory\_Orya}, \texttt{sat\_Beng}, \texttt{taq\_Latn}, \texttt{taq\_Tfng}, \texttt{tzm\_Tfng}) lack a dedicated model due to missing script coverage or absence from Goldfish entirely.

\begin{table*}[t]
\centering
\scriptsize
\setlength{\tabcolsep}{3pt}
\begin{tabular}{@{}llllrcccccc@{}}
\toprule
 & & & & & \multicolumn{4}{c}{Model} & \\
\cmidrule(lr){6-9}
Language & FLORES & Label & Family & Res. & Goldfish & BLOOM & XGLM & mGPT & PUD \\
\midrule
Afrikaans & afr\_Latn & NoDom-RRRLR & Indo-European & 3 & \checkmark & $\times$ & $\times$ & \checkmark & $\times$ \\
Akan & aka\_Latn & SVO-RNLR? & Niger-Congo & 1 & $\times$ & \checkmark$^{*}$ & $\times$ & $\times$ & $\times$ \\
Amharic & amh\_Ethi & SOV-LLRLL & Afro-Asiatic & 2 & \checkmark & $\times$ & $\times$ & $\times$ & $\times$ \\
Armenian & hye\_Armn & NoDom-RRLLN & Indo-European & 1 & \checkmark & $\times$ & $\times$ & \checkmark & $\times$ \\
Assamese & asm\_Beng & SOV-L??L? & Indo-European & 1 & $\times$ & \checkmark$^{*}$ & $\times$ & $\times$ & $\times$ \\
Ayacucho Quechua & quy\_Latn & SOV-L?LLL & Quechuan & -- & $\times$ & $\times$ & \checkmark$^{*}$ & $\times$ & $\times$ \\
Bambara & bam\_Latn & SOV-LLLRL & Niger-Congo & 1 & $\times$ & \checkmark & $\times$ & $\times$ & $\times$ \\
Bashkir & bak\_Cyrl & SOV-LRLLL & Turkic & 1 & $\times$ & $\times$ & $\times$ & \checkmark & $\times$ \\
Basque & eus\_Latn & SOV-LLLRL & Isolate & 4 & \checkmark & \checkmark & \checkmark & \checkmark & $\times$ \\
Belarusian & bel\_Cyrl & SVO-RRRLR & Indo-European & 3 & \checkmark & $\times$ & $\times$ & \checkmark & $\times$ \\
Bengali & ben\_Beng & SOV-LRLLL & Indo-European & 3 & \checkmark & \checkmark & \checkmark & \checkmark & $\times$ \\
Bosnian & bos\_Latn & SVO-RRRLR & Indo-European & 3 & \checkmark & $\times$ & $\times$ & $\times$ & $\times$ \\
Bulgarian & bul\_Cyrl & SVO-RRRLR & Indo-European & 3 & \checkmark & $\times$ & \checkmark & \checkmark & $\times$ \\
Burmese & mya\_Mymr & SOV-LNLRL & Sino-Tibetan & 1 & $\times$ & $\times$ & \checkmark & \checkmark & $\times$ \\
Catalan & cat\_Latn & SVO-RRRRR & Indo-European & 4 & \checkmark & \checkmark & \checkmark & $\times$ & $\times$ \\
Chinese (Simp.) & zho\_Hans & SVO-RRRLL & Sino-Tibetan & 5 & \checkmark & \checkmark & \checkmark & $\times$ & \checkmark \\
Chinese (Trad.) & zho\_Hant & SVO-RRRLL & Sino-Tibetan & 5 & $\times$ & \checkmark & $\times$ & $\times$ & $\times$ \\
Croatian & hrv\_Latn & SVO-RRRLR & Indo-European & 4 & \checkmark & $\times$ & $\times$ & $\times$ & $\times$ \\
Czech & ces\_Latn & SVO-RRRLR & Indo-European & 4 & \checkmark & $\times$ & $\times$ & $\times$ & \checkmark \\
Danish & dan\_Latn & SVO-RRRLR & Indo-European & 3 & \checkmark & $\times$ & $\times$ & \checkmark & $\times$ \\
Dutch & nld\_Latn & NoDom-RRRLR & Indo-European & 4 & \checkmark & $\times$ & $\times$ & $\times$ & $\times$ \\
Eastern Panjabi & pan\_Guru & SOV-RRLLN & Indo-European & 2 & \checkmark & \checkmark & $\times$ & $\times$ & $\times$ \\
English & eng\_Latn & SVO-RRRLR & Indo-European & 5 & \checkmark & \checkmark & \checkmark & \checkmark & \checkmark \\
Esperanto & epo\_Latn & NoDom-RRRLR & Constructed & 1 & \checkmark & $\times$ & $\times$ & $\times$ & $\times$ \\
Estonian & est\_Latn & SVO-RRLLR & Uralic & 3 & \checkmark & $\times$ & \checkmark & \checkmark & $\times$ \\
Finnish & fin\_Latn & SVO-RRLLR & Uralic & 4 & \checkmark & $\times$ & \checkmark & \checkmark & \checkmark \\
Fon & fon\_Latn & SVO-RRNRR & Niger-Congo & 0 & $\times$ & \checkmark & $\times$ & $\times$ & $\times$ \\
French & fra\_Latn & SVO-RRRRR & Indo-European & 5 & \checkmark & \checkmark & \checkmark & \checkmark & \checkmark \\
Galician & glg\_Latn & NoDom-RRRRR & Indo-European & 3 & \checkmark & $\times$ & $\times$ & $\times$ & $\times$ \\
Ganda & lug\_Latn & SVO-RRRRR & Niger-Congo & 1 & $\times$ & \checkmark & $\times$ & $\times$ & $\times$ \\
Georgian & kat\_Geor & NoDom-RRLLR & Kartvelian & 3 & \checkmark & $\times$ & $\times$ & \checkmark & $\times$ \\
German & deu\_Latn & NoDom-RRRLR & Indo-European & 5 & \checkmark & $\times$ & \checkmark & \checkmark & \checkmark \\
Greek & ell\_Grek & NoDom-RRRLR & Indo-European & 3 & \checkmark & $\times$ & \checkmark & \checkmark & $\times$ \\
Gujarati & guj\_Gujr & SOV-LRLLL & Indo-European & 1 & \checkmark & \checkmark & $\times$ & $\times$ & $\times$ \\
Haitian Creole & hat\_Latn & SVO-RRRRR & Creole & 0 & $\times$ & $\times$ & \checkmark & $\times$ & $\times$ \\
Halh Mongolian & khk\_Cyrl & SOV-LLLLL & Mongolic & -- & $\times$ & $\times$ & $\times$ & \checkmark & $\times$ \\
Hausa & hau\_Latn & SVO-RRRLR & Afro-Asiatic & 2 & \checkmark & $\times$ & $\times$ & $\times$ & $\times$ \\
Hebrew & heb\_Hebr & SVO-RRRRR & Afro-Asiatic & 3 & \checkmark & $\times$ & $\times$ & \checkmark & $\times$ \\
Hindi & hin\_Deva & SOV-LRLLL & Indo-European & 4 & \checkmark & \checkmark & \checkmark & \checkmark & \checkmark \\
Hungarian & hun\_Latn & NoDom-RRLLN & Uralic & 4 & \checkmark & $\times$ & $\times$ & \checkmark & $\times$ \\
Icelandic & isl\_Latn & SVO-RRRLR & Indo-European & 2 & \checkmark & $\times$ & $\times$ & $\times$ & \checkmark \\
Igbo & ibo\_Latn & SVO-RRRRR & Niger-Congo & 1 & $\times$ & \checkmark & $\times$ & $\times$ & $\times$ \\
Indonesian & ind\_Latn & SVO-RRRRR & Austronesian & 3 & \checkmark & \checkmark & \checkmark & \checkmark & \checkmark \\
Italian & ita\_Latn & NoDom-RRRRR & Indo-European & 4 & \checkmark & $\times$ & \checkmark & \checkmark & \checkmark \\
Japanese & jpn\_Jpan & SOV-LLLLL & Japonic & 5 & \checkmark & $\times$ & \checkmark & \checkmark & \checkmark \\
Javanese & jav\_Latn & SVO-RRRRR & Austronesian & 1 & $\times$ & $\times$ & $\times$ & \checkmark & $\times$ \\
Kannada & kan\_Knda & SOV-LLLLL & Dravidian & 1 & \checkmark & \checkmark & $\times$ & $\times$ & $\times$ \\
Kazakh & kaz\_Cyrl & SOV-L???? & Turkic & 3 & \checkmark$^{*}$ & $\times$ & $\times$ & \checkmark$^{*}$ & $\times$ \\
Kikuyu & kik\_Latn & SVO-RRRRR & Niger-Congo & 1 & $\times$ & \checkmark & $\times$ & $\times$ & $\times$ \\
Kinyarwanda & kin\_Latn & SVO-R?RR? & Niger-Congo & 1 & $\times$ & \checkmark$^{*}$ & $\times$ & $\times$ & $\times$ \\
Korean & kor\_Hang & SOV-LLLLL & Koreanic & 4 & \checkmark & $\times$ & \checkmark & \checkmark & \checkmark \\
Kyrgyz & kir\_Cyrl & SOV-L?LL? & Turkic & 1 & \checkmark$^{*}$ & $\times$ & $\times$ & \checkmark$^{*}$ & $\times$ \\
\bottomrule
\end{tabular}
\caption{Evaluation languages (part 1 of 2): the 103 FLORES-200 
languages with a base word order label appearing in at least one 
evaluation set. \emph{Label}: dominant order + five head-direction 
parameters following the encoding in 
Figure~\ref{fig:switchable-rules} (NoDom = No Dominant Order). 
\emph{Res.}: resource class 0--5 \citep{joshi-etal-2020-state}; 
-- = not classified. \emph{Model} columns: presence in training 
data (Goldfish = available at all four training sizes, 68/103; 
BLOOM, 46; XGLM, 30; mGPT, 56). \emph{PUD}: PUD treebank 
available (20/103; all PUD languages fall within this set). 
\checkmark$^{*}$: included for base word order analysis but 
excluded from directionality analysis (\S\ref{sec:results}) due to missing 
parameters~(\texttt{?}). N = no dominant order.}
\label{tab:multilingual_languages}
\end{table*}

\begin{table*}[t]
\centering
\scriptsize
\setlength{\tabcolsep}{3pt}
\begin{tabular}{@{}llllrcccccc@{}}
\toprule
 & & & & & \multicolumn{4}{c}{Model} & \\
\cmidrule(lr){6-9}
Language & FLORES & Label & Family & Res. & Goldfish & BLOOM & XGLM & mGPT & PUD \\
\midrule
Lingala & lin\_Latn & SVO-R?RRR & Niger-Congo & 1 & $\times$ & \checkmark$^{*}$ & $\times$ & $\times$ & $\times$ \\
Lithuanian & lit\_Latn & NoDom-RRRLR & Indo-European & 3 & \checkmark & $\times$ & $\times$ & \checkmark & $\times$ \\
Macedonian & mkd\_Cyrl & SVO-RRRLR & Indo-European & 1 & \checkmark & $\times$ & $\times$ & $\times$ & $\times$ \\
Malayalam & mal\_Mlym & SOV-LLLLL & Dravidian & 1 & \checkmark & \checkmark & $\times$ & \checkmark & $\times$ \\
Maltese & mlt\_Latn & SVO-RRRRR & Afro-Asiatic & 2 & \checkmark & $\times$ & $\times$ & $\times$ & $\times$ \\
Marathi & mar\_Deva & SOV-RNLLL & Indo-European & 2 & \checkmark & \checkmark & $\times$ & \checkmark & $\times$ \\
MSA & arb\_Arab & VSO-RRRRR & Afro-Asiatic & -- & \checkmark & \checkmark & \checkmark & \checkmark & \checkmark \\
Nepali & npi\_Deva & SOV-LLLL? & Indo-European & 1 & $\times$ & \checkmark$^{*}$ & $\times$ & $\times$ & $\times$ \\
Northern Sotho & nso\_Latn & SVO-R???? & Niger-Congo & 1 & $\times$ & \checkmark$^{*}$ & $\times$ & $\times$ & $\times$ \\
Northern Uzbek & uzn\_Latn & SOV-LRLLL & Turkic & -- & $\times$ & $\times$ & $\times$ & \checkmark & $\times$ \\
Norwegian Bokm\aa l & nob\_Latn & SVO-RRRLR & Indo-European & 1 & \checkmark & $\times$ & $\times$ & $\times$ & $\times$ \\
Nyanja & nya\_Latn & SVO-RRRRR & Niger-Congo & 1 & $\times$ & \checkmark & $\times$ & $\times$ & $\times$ \\
Odia & ory\_Orya & SOV-?NLLN & Indo-European & 1 & $\times$ & \checkmark$^{*}$ & $\times$ & $\times$ & $\times$ \\
Polish & pol\_Latn & SVO-RRRLR & Indo-European & 4 & \checkmark & $\times$ & $\times$ & \checkmark & \checkmark \\
Portuguese & por\_Latn & SVO-RRRRR & Indo-European & 4 & \checkmark & \checkmark & \checkmark & \checkmark & \checkmark \\
Romanian & ron\_Latn & SVO-RRRRR & Indo-European & 3 & \checkmark & $\times$ & $\times$ & \checkmark & $\times$ \\
Rundi & run\_Latn & SVO-RRRRR & Niger-Congo & 0 & $\times$ & \checkmark & $\times$ & $\times$ & $\times$ \\
Russian & rus\_Cyrl & SVO-RRRLR & Indo-European & 4 & \checkmark & $\times$ & \checkmark & \checkmark & \checkmark \\
Serbian & srp\_Cyrl & SVO-RRRLR & Indo-European & 4 & \checkmark & $\times$ & $\times$ & $\times$ & $\times$ \\
Shona & sna\_Latn & SVO-RRRRR & Niger-Congo & 1 & $\times$ & \checkmark & $\times$ & $\times$ & $\times$ \\
Sinhala & sin\_Sinh & SOV-LLLLL & Indo-European & 0 & \checkmark & $\times$ & $\times$ & $\times$ & $\times$ \\
Slovak & slk\_Latn & SVO-RRRLR & Indo-European & 3 & \checkmark & $\times$ & $\times$ & $\times$ & $\times$ \\
Slovenian & slv\_Latn & SVO-RRRLR & Indo-European & 3 & \checkmark & $\times$ & $\times$ & $\times$ & $\times$ \\
Somali & som\_Latn & SOV-LRNRR & Afro-Asiatic & 1 & \checkmark & $\times$ & $\times$ & $\times$ & $\times$ \\
Southern Sotho & sot\_Latn & SVO-RRRRR & Niger-Congo & 1 & $\times$ & \checkmark & $\times$ & $\times$ & $\times$ \\
Spanish & spa\_Latn & SVO-RRRRR & Indo-European & 5 & \checkmark & \checkmark & \checkmark & \checkmark & \checkmark \\
Std.\ Latvian & lvs\_Latn & SVO-RRRLR & Indo-European & -- & $\times$ & $\times$ & $\times$ & \checkmark & $\times$ \\
Std.\ Malay & zsm\_Latn & SVO-RRRRR & Austronesian & -- & $\times$ & $\times$ & $\times$ & \checkmark & $\times$ \\
Swahili & swh\_Latn & SVO-RRRRR & Niger-Congo & -- & \checkmark & \checkmark & \checkmark & \checkmark & $\times$ \\
Swedish & swe\_Latn & SVO-RRRLR & Indo-European & 4 & \checkmark & $\times$ & $\times$ & \checkmark & \checkmark \\
Tagalog & tgl\_Latn & VSO-RRRNR & Austronesian & 3 & \checkmark & $\times$ & $\times$ & $\times$ & $\times$ \\
Tajik & tgk\_Cyrl & SOV-LRRRR & Indo-European & 1 & \checkmark & $\times$ & $\times$ & $\times$ & $\times$ \\
Tamil & tam\_Taml & SOV-LLLLL & Dravidian & 3 & \checkmark & \checkmark & \checkmark & $\times$ & $\times$ \\
Tatar & tat\_Cyrl & SOV-LRLLL & Turkic & 1 & \checkmark & $\times$ & $\times$ & \checkmark & $\times$ \\
Telugu & tel\_Telu & SOV-LLLLL & Dravidian & 1 & \checkmark & \checkmark & \checkmark & \checkmark & $\times$ \\
Thai & tha\_Thai & SVO-RRRRR & Tai-Kadai & 3 & \checkmark & $\times$ & \checkmark & \checkmark & \checkmark \\
Tsonga & tso\_Latn & SVO-RR?R? & Niger-Congo & 1 & $\times$ & \checkmark$^{*}$ & $\times$ & $\times$ & $\times$ \\
Tswana & tsn\_Latn & SVO-RRRRR & Niger-Congo & 2 & $\times$ & \checkmark & $\times$ & $\times$ & $\times$ \\
Tumbuka & tum\_Latn & SVO-R???? & Niger-Congo & 1 & $\times$ & \checkmark$^{*}$ & $\times$ & $\times$ & $\times$ \\
Turkish & tur\_Latn & SOV-LLLLL & Turkic & 4 & \checkmark & $\times$ & \checkmark & \checkmark & \checkmark \\
Turkmen & tuk\_Latn & SOV-LLLLL & Turkic & 1 & $\times$ & $\times$ & $\times$ & \checkmark & $\times$ \\
Twi & twi\_Latn & SVO-RNLR? & Niger-Congo & 1 & $\times$ & \checkmark$^{*}$ & $\times$ & $\times$ & $\times$ \\
Ukrainian & ukr\_Cyrl & SVO-RRRLR & Indo-European & 3 & \checkmark & $\times$ & $\times$ & \checkmark & $\times$ \\
Urdu & urd\_Arab & SOV-LRLLL & Indo-European & 3 & \checkmark & \checkmark & \checkmark & $\times$ & $\times$ \\
Vietnamese & vie\_Latn & SVO-RRRRR & Austroasiatic & 4 & \checkmark & \checkmark & \checkmark & \checkmark & $\times$ \\
Welsh & cym\_Latn & VSO-RRRRR & Indo-European & 1 & \checkmark & $\times$ & $\times$ & $\times$ & $\times$ \\
Western Persian & pes\_Arab & SOV-?RRRR & Indo-European & -- & \checkmark$^{*}$ & $\times$ & $\times$ & \checkmark$^{*}$ & $\times$ \\
Wolof & wol\_Latn & SVO-RRRRR & Niger-Congo & 2 & $\times$ & \checkmark & $\times$ & $\times$ & $\times$ \\
Xhosa & xho\_Latn & SVO-RRRRR & Niger-Congo & 2 & $\times$ & \checkmark & $\times$ & $\times$ & $\times$ \\
Yoruba & yor\_Latn & SVO-RRRRR & Niger-Congo & 2 & $\times$ & \checkmark & $\times$ & \checkmark & $\times$ \\
Zulu & zul\_Latn & SVO-RRRRR & Niger-Congo & 2 & $\times$ & \checkmark & $\times$ & $\times$ & $\times$ \\
\bottomrule
\end{tabular}
\caption{Evaluation languages (part 2 of 2, continued from 
Table~\ref{tab:multilingual_languages}). Column definitions are 
identical. MSA = Modern Standard Arabic.}
\label{tab:multilingual_languages_2}
\end{table*}

\subsection{Evaluation}
\label{app:eval-metrics}

\paragraph{BPEC}
\label{app:bpec}
For artificial languages, raw perplexity is directly comparable because all 192 variants share the same vocabulary, tokeniser, and derivation probabilities. For natural languages, per-token PPL is not comparable across scripts and tokenisers. We therefore follow \citet{cotterell-bpec} and normalise corpus negative log-likelihood by the English character count of the parallel corpus to obtain bits per English character (BPEC):
\begin{equation}
\mathrm{BPEC}_{\ell}
= \frac{-\displaystyle\sum_{s \in S_{\ell}}\;\sum_{t=1}^{n_s}
        \log_2 p_{\theta}\!\bigl(x_t^{(s)} \mid x_{<t}^{(s)}\bigr)}
       {C_{\mathrm{eng}}}
\end{equation}
where $S_\ell$ are the evaluation sentences in language~$\ell$, 
$n_s$ the scored subword tokens per sentence, and 
$C_{\mathrm{eng}}$ the total character count of the English 
portion of the respective corpus (FLORES-200/PUD). We prepend 
the model's start token (\texttt{[CLS]} for Goldfish, 
\texttt{[BOS]} for multilingual models) as conditioning context, 
but exclude it from the scored tokens.

\paragraph{Greenbergian Universals}
\label{app:universals}
From Greenberg's 45 universals \citep{greenberg1963universals}, 
we select the subset testable with syntactic constructions 
in our artificial languages. Since our PCFG generates only 
declarative sentences, we are limited to universals involving 
base S/V/O order, adpositions, adjective--noun order, and 
relative clauses. This yields four implicational universals: 
Universal~1 (subject precedes object in the dominant orders SOV, 
SVO, VSO), Universal~3 (VSO languages are prepositional), 
Universal~4 (SOV languages are postpositional), and Universal~17 
(VSO languages place the adjective after the noun). We 
also test Dryer's observation that OV languages correlate 
with prenominal relative clauses and VO languages with 
postnominal ones \citep{dryer1992greenbergian}. For each 
universal, we compare mean test PPL between the 
configuration predicted by the universal and its counterpart.

\subsection{Language Resource Level and Word Order Distribution}
\label{app:resource-word-order}

We use the resource taxonomy of \citet{joshi-etal-2020-state}, which 
classifies languages into six classes (0--5) based on the 
availability of labeled and unlabeled NLP data, as a proxy for 
resourcedness (column \emph{Res.}\ in 
Tables~\ref{tab:multilingual_languages}--\ref{tab:multilingual_languages_2}). 
We report statistics on the 68 Goldfish languages listed in
Tables~\ref{tab:multilingual_languages}--\ref{tab:multilingual_languages_2},
tracked across all four training scales, as these are the 
languages for which we directly observe the emerging SVO 
advantage (65 with a resource class assignment in \citet{joshi-etal-2020-state}). SVO languages have substantially higher 
resourcedness than SOV (mean resource class 3.35 vs.\ 2.24; 
Mann--Whitney $U{=}472$, $p{=}0.003$, $r{=}0.72$). Of the 19 
highly-resourced (Class~4--5) \{SVO, SOV\} languages, 14 are SVO 
(74\%), and a Cochran--Armitage trend test confirms that the 
proportion of SVO rises monotonically with resource class 
($z{=}3.03$, $p{=}0.002$).


\subsection{Mixed-Effects Analysis of the SVO--SOV Gap}
\label{app:mixed-effects}
In line with prior work that uses mixed-effects models to separate language-modeling difficulty from confounding language properties \citep{mielke-etal-2019-kind}, we fit linear mixed-effects models to the Goldfish BPEC of the 50 SVO/SOV languages (32 SVO, 18 SOV) in our full directionality-annotated set (Figure~\ref{fig:rightness-score}c,d), across all four training sizes (200 language-by-size observations; Table~\ref{tab:mixed-effects}); VSO and NoDominant languages also shown in Figure~\ref{fig:rightness-score}c,d are excluded, as the model contrasts SVO against SOV. The model is
\begin{equation*}
\begin{aligned}
\mathtt{bpec} \sim{}& \mathtt{is\_SOV}\times\mathtt{size} + \mathrm{covariates} \\
&{}+ (1\mid\mathtt{language}),
\end{aligned}
\end{equation*}
with training size entered as z-scored $\log_{10}$(MB), so main effects are read at the centre of the size range ($\sim$47\,MB) and interactions per SD of log-size ($\sim$8$\times$ data). The covariates (all reported in Table~\ref{tab:mixed-effects}) are morphological complexity (\texttt{mattr\_z}), language resourcedness (\texttt{resource\_lvl}, z-scored resourcedness class from \citealp{joshi-etal-2020-state}), and training data composition (\texttt{data\_comp}, OSCAR web-crawl share).\footnote{The share of each language's Goldfish training data drawn from OSCAR rather than more curated corpora, computed from the per-language \texttt{proportions} column of the Goldfish data documentation: \url{https://github.com/tylerachang/goldfish/blob/main/data/goldfish_data_info.tsv}.} We measure morphological complexity as subword MATTR (1000-token window) under each language's own tokeniser \citep[as in][]{tatariya-etal-2025-interplay}, since corpus-based measures such as TTR have been shown to correlate with, and be interchangeable with, typological complexity measures derived from databases such as WALS \citep{bentz-etal-2016-comparison}, and it is available for all languages, so it does not limit the sample. Models are fit by maximum likelihood so the nested models M1--M9 are comparable; the robustness block in Table~\ref{tab:mixed-effects} refits key models with a maximal random-effects structure (random intercept + random size slope per language). Our finding is the \texttt{is\_SOV\,$\times$\,size} interaction (the growth of the SVO--SOV performance gap with training data scale), and each covariate is tested as a predictor of it, both as a level term and as an interaction. Table~\ref{tab:mixed-effects} reports all coefficients across the nine nested models (M1--M9) and the robustness checks.
\begin{table}[t!]
\centering
\footnotesize
\setlength{\tabcolsep}{4pt}
\begin{tabular}{@{}lrrr@{}}
\toprule
Term & $\beta$ & SE & $p$ \\
\midrule
\multicolumn{4}{@{}l}{\textit{M1 base (no covariates)}} \\
\texttt{is\_SOV}            & \textbf{+0.071} & 0.024 & \textbf{.003} \\
\texttt{is\_SOV$\times$size} & \textbf{+0.049} & 0.011 & \textbf{$<$.001} \\
\multicolumn{4}{@{}l}{\textit{M2 + morphological complexity (level)}} \\
\texttt{is\_SOV}            & \textbf{+0.050} & 0.024 & \textbf{.041} \\
\texttt{is\_SOV$\times$size} & \textbf{+0.049} & 0.011 & \textbf{$<$.001} \\
\texttt{mattr\_z}           & \textbf{+0.027} & 0.012 & \textbf{.021} \\
\multicolumn{4}{@{}l}{\textit{M3 + language resourcedness (level)}} \\
\texttt{is\_SOV}            & \textbf{+0.053} & 0.025 & \textbf{.033} \\
\texttt{is\_SOV$\times$size} & \textbf{+0.049} & 0.011 & \textbf{$<$.001} \\
\texttt{resource\_lvl}      & $-0.023$ & 0.012 & .057 \\
\multicolumn{4}{@{}l}{\textit{M4 + training data composition (level)}} \\
\texttt{is\_SOV}            & \textbf{+0.052} & 0.025 & \textbf{.034} \\
\texttt{is\_SOV$\times$size} & \textbf{+0.049} & 0.011 & \textbf{$<$.001} \\
\texttt{data\_comp}         & $-0.021$ & 0.012 & .076 \\
\multicolumn{4}{@{}l}{\textit{M5 + all three levels}} \\
\texttt{is\_SOV}            & +0.030 & 0.025 & .224 \\
\texttt{is\_SOV$\times$size} & \textbf{+0.049} & 0.011 & \textbf{$<$.001} \\
\texttt{mattr\_z}           & \textbf{+0.025} & 0.011 & \textbf{.022} \\
\texttt{resource\_lvl}      & $-0.011$ & 0.014 & .441 \\
\texttt{data\_comp}         & $-0.014$ & 0.014 & .317 \\
\multicolumn{4}{@{}l}{\textit{M6 + morphological complexity $\times$ word order}} \\
\texttt{is\_SOV}            & \textbf{+0.055} & 0.025 & \textbf{.029} \\
\texttt{is\_SOV$\times$size} & \textbf{+0.049} & 0.011 & \textbf{$<$.001} \\
\texttt{mattr\_z}           & \textbf{+0.034} & 0.014 & \textbf{.018} \\
\texttt{mattr\_z$\times$is\_SOV} & $-0.022$ & 0.025 & .386 \\
\multicolumn{4}{@{}l}{\textit{M7 + language resourcedness $\times$ size}} \\
\texttt{is\_SOV}            & +0.036 & 0.025 & .147 \\
\texttt{is\_SOV$\times$size} & \textbf{+0.039} & 0.012 & \textbf{$<$.001} \\
\texttt{resource\_lvl$\times$size} & \textbf{$-0.012$} & 0.006 & \textbf{.037} \\
\texttt{mattr\_z}           & \textbf{+0.024} & 0.012 & \textbf{.037} \\
\multicolumn{4}{@{}l}{\textit{M8 + training data composition $\times$ size}} \\
\texttt{is\_SOV}            & +0.032 & 0.025 & .202 \\
\texttt{is\_SOV$\times$size} & \textbf{+0.040} & 0.012 & \textbf{$<$.001} \\
\texttt{data\_comp}         & $-0.021$ & 0.011 & .064 \\
\texttt{data\_comp$\times$size} & $-0.009$ & 0.006 & .098 \\
\texttt{mattr\_z}           & \textbf{+0.027} & 0.011 & \textbf{.015} \\
\multicolumn{4}{@{}l}{\textit{M9 full model (resourcedness \& composition $\times$ size)}} \\
\texttt{is\_SOV}            & +0.030 & 0.025 & .224 \\
\texttt{is\_SOV$\times$size} & \textbf{+0.038} & 0.012 & \textbf{.001} \\
\texttt{mattr\_z}           & \textbf{+0.025} & 0.011 & \textbf{.022} \\
\texttt{resource\_lvl$\times$size} & $-0.010$ & 0.007 & .165 \\
\texttt{data\_comp$\times$size} & $-0.003$ & 0.007 & .634 \\
\multicolumn{4}{@{}l}{\textit{Robustness: maximal random effects}} \\
M1: \texttt{is\_SOV$\times$size} & \textbf{+0.049} & 0.011 & \textbf{$<$.001} \\
M7: \texttt{is\_SOV$\times$size} & \textbf{+0.039} & 0.012 & \textbf{.001} \\
M9: \texttt{is\_SOV$\times$size} & \textbf{+0.038} & 0.012 & \textbf{.002} \\
binary res.\ \texttt{is\_SOV$\times$size} & \textbf{+0.037} & 0.011 & \textbf{.001} \\
binary res.\ \texttt{high\_res$\times$size} & \textbf{$-0.032$} & 0.012 & \textbf{.007} \\
\bottomrule
\end{tabular}
\caption{Mixed-effects models of Goldfish BPEC (50 SVO/SOV languages $\times$ 4 sizes). \textbf{Bold}\,=\,$p<.05$. Each covariate (morphological complexity, resourcedness, and training-data composition, OSCAR share) is entered as a level term (M2--M4), jointly (M5), and in interaction (M6--M9). Our finding, the word order $\times$ size interaction (\texttt{is\_SOV$\times$size}, the SOV gap growing with data), survives every control (M2--M9) and a maximal random-effects structure. The main effect (\texttt{is\_SOV}, the SVO--SOV gap at the average training size, $\sim$47\,MB) is absorbed once all three covariate levels are added together (M5).}
\label{tab:mixed-effects}
\end{table}
Our finding, the \texttt{is\_SOV$\times$size} interaction, survives every control: it stays significant across M2--M9 and under the maximal random-effects structure, shrinking only ${\sim}$20\% (from ${+}0.049$ to ${+}0.038$) when resourcedness is allowed to scale with data (M7, M9). The main effect (\texttt{is\_SOV}, the average SVO--SOV gap), by contrast, is absorbed: it is significant on its own (M1) and with any single covariate (M2--M4) but non-significant once all three are added together (M5), so the covariates jointly explain the average gap, but not its growth with data. Among the covariates, only resourcedness has a significant standalone size interaction (\texttt{resource\_lvl$\times$size}, $\beta{=}{-}0.012$, $p{=}.037$; M7), in the same direction as our effect: higher-resource languages improve faster with data, widening the gap for the lower-resourced SOV group, which accounts for the ${\sim}$20\% reduction. Morphological complexity does not act differently on SOV (\texttt{mattr\_z$\times$is\_SOV}, $p{=}.39$; M6) and its own effect shrinks with data (within SVO, MATTR $\times$ log-size $\beta{=}{-}0.017$, $p{=}.014$), so it cannot explain an effect that grows with data. Training data composition shows no size interaction on its own (\texttt{data\_comp$\times$size}, $p{=}.098$; M8); in the full model (M9) it is collinear with resourcedness ($r{=}0.64$), so neither size interaction is individually significant there, though the word order interaction is unchanged.

\subsection{Additional Results}
\label{app:additional-results}

\begin{figure}[t]
  \centering
  \includegraphics[width=\columnwidth]{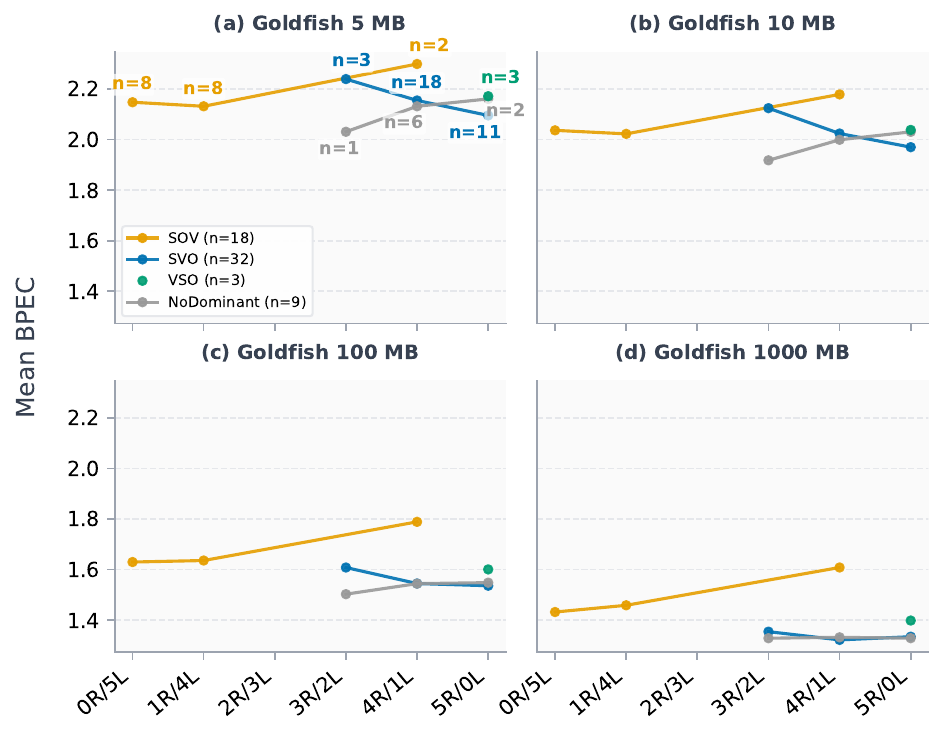}
\caption{Directionality vs.\ BPEC at all four Goldfish scales, same 62 languages (cf.\ Figure~\ref{fig:rightness-score}c--d for 5 and 1\,GB only). SVO separates from SOV at 100\,MB--1\,GB.}
  \label{fig:rightness-goldfish}
\end{figure}

\begin{figure}[t]
  \centering
  \includegraphics[width=\linewidth]{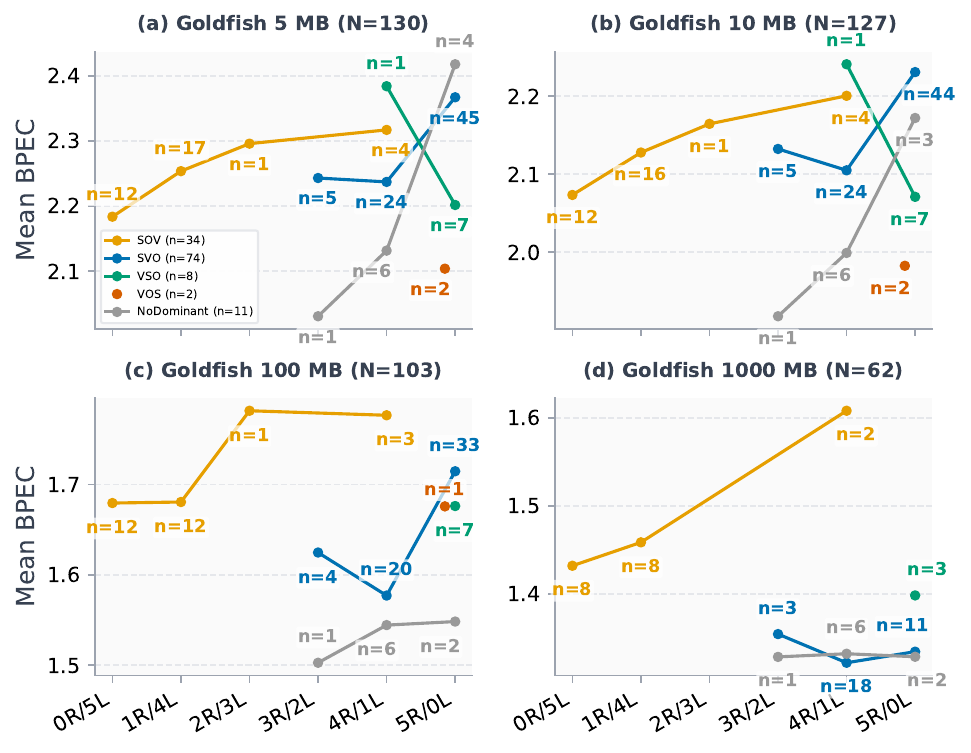}
\caption{As Figure~\ref{fig:rightness-goldfish} but with all available languages per scale. Same pattern, ruling out selection artefacts.}
  \label{fig:goldfish_all_langs_directionality}
\end{figure}

\begin{figure}[t]
\centering
\includegraphics[width=\columnwidth]{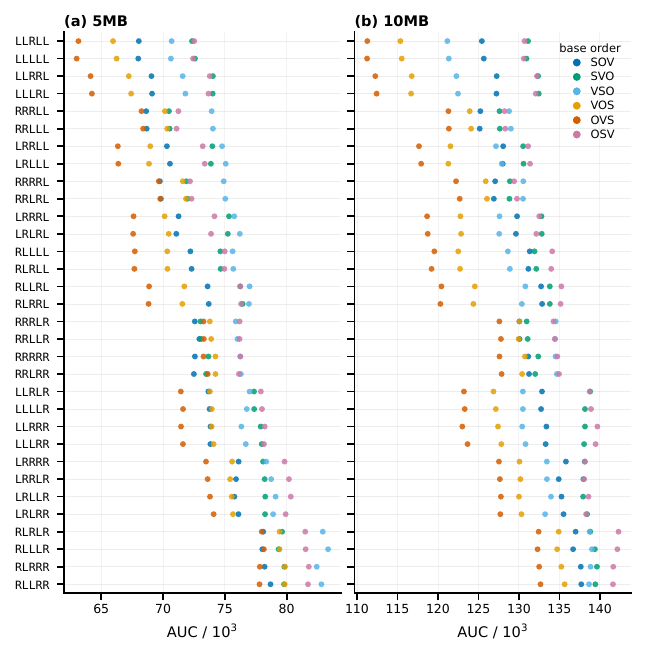}
\caption{Convergence AUC ($\div 10^{3}$; lower = faster)
for all 32 artificial language configurations at 5\,MB~(a) and 10\,MB~(b),
ranked by cross-order mean. Rankings stable across
sizes (Spearman $\rho = 0.884$) and strongly correlated
with final PPL ($\rho > 0.89$).}
\label{fig:convergence-ranking}
\end{figure}

\begin{figure*}[t!]
  \centering
  \includegraphics[width=\textwidth]{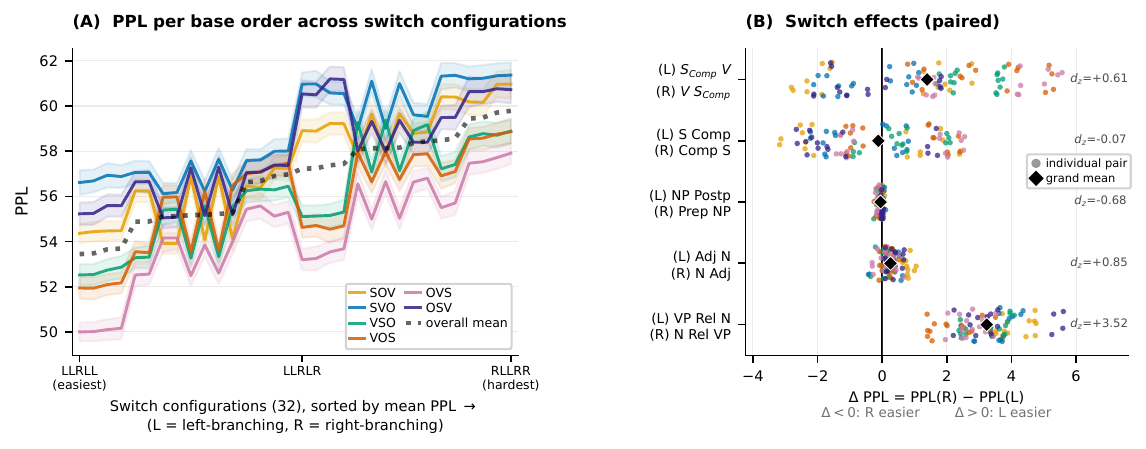}
\caption{
Replication of \autoref{fig:fig-artificial-5mb} with 10\,MB
training samples. Setup and notation are identical.
All five paired $t$-tests remain significant after
Holm--Bonferroni correction ($p<.001$).
}
  \label{fig:fig-artificial-10mb}
\end{figure*}

\begin{figure*}[t]
    \centering
    \includegraphics[width=\textwidth]{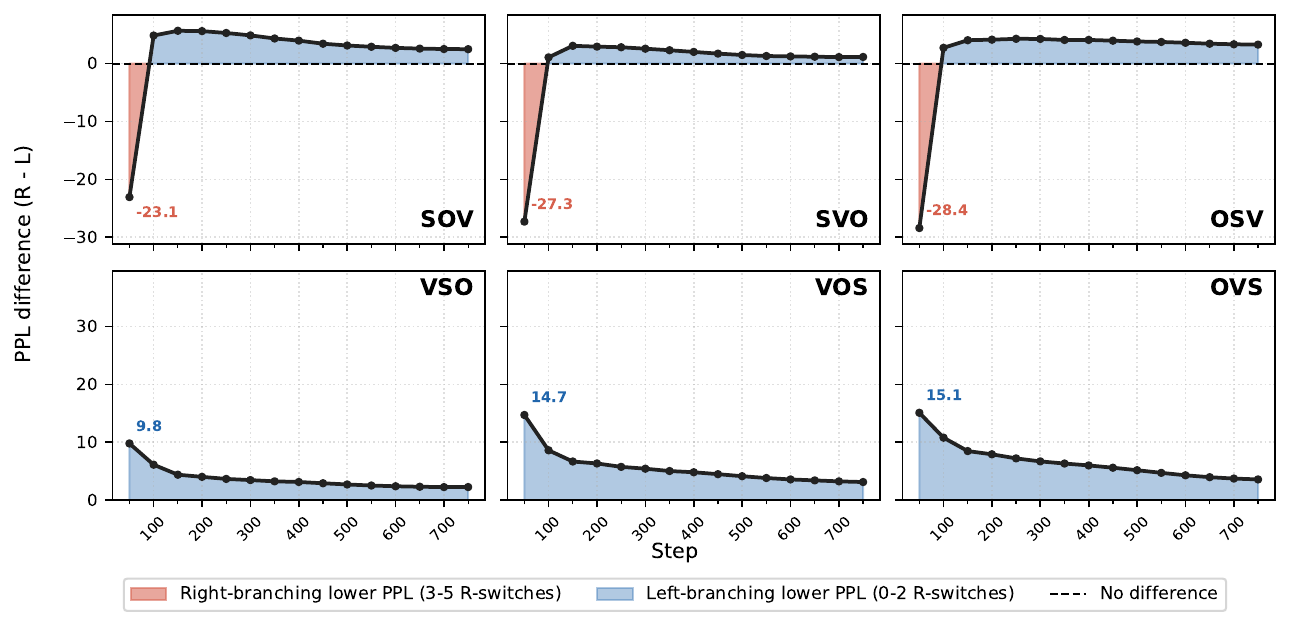}
    \caption{Extended version of Figure~\ref{fig:ppl_diff_2panel}
showing all six base orders. SV orders:
right-branching initially easier, flips after ${\sim}$100 steps.
VS orders: left-branching preferred throughout.
Same trend at 10\,MB.}
    \label{fig:ppl_diff_5mb}
\end{figure*}

\begin{figure}[t]
  \centering
  \includegraphics[width=\linewidth]{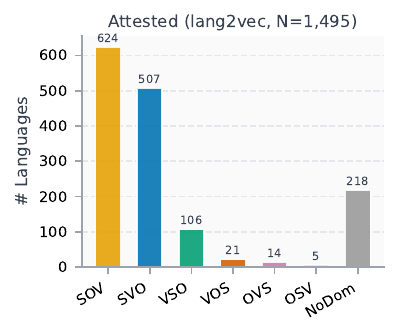}
\caption{Base word order distribution across attested languages. Figure~\ref{fig:rightness-score}a uses only the $N{=}526$ subset with complete switch annotations for all five parameters, which skews toward well-documented, right-branching Indo-European languages.}
  \label{fig:combined-encoding-basewo}
\end{figure}

\begin{figure*}[t]
    \centering
    \includegraphics[width=\textwidth]{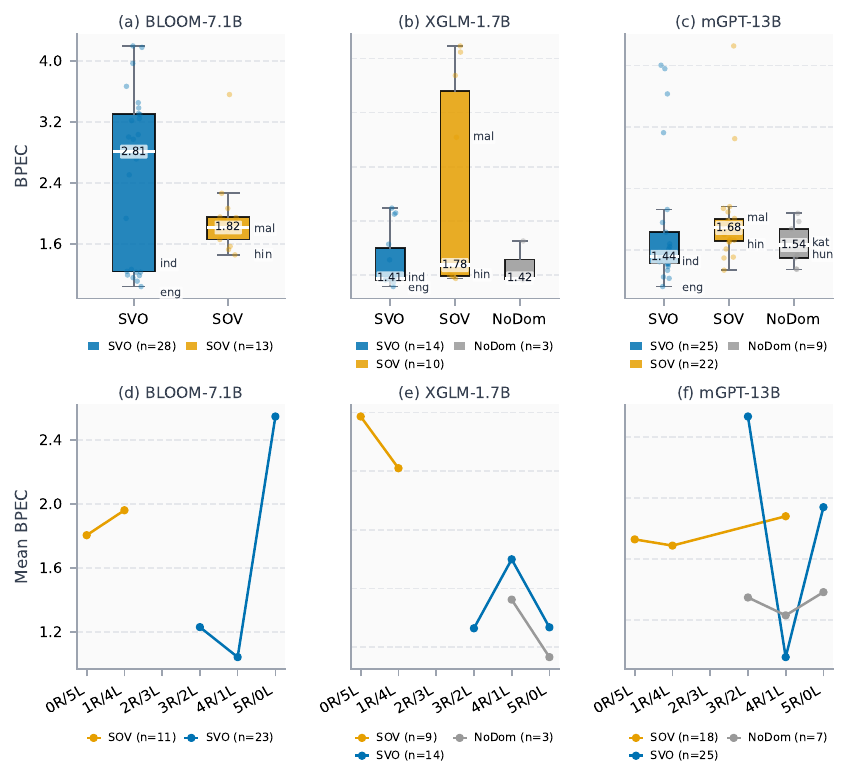}
\caption{BPEC by base word order (a--c) and head directionality (d--f) for the largest model in each multilingual family (FLORES-200); smaller sizes show the same patterns. SVO compresses better than SOV in XGLM and mGPT; BLOOM reverses because 21 of 28 SVO languages are very-low-resource Niger-Congo languages (\S\ref{sec:natural_results}). The right-branching SVO advantage is visible even in BLOOM's directionality panels (d--f), where it is masked at the base word order level.}
    \label{fig:multilingual_combined}
\end{figure*}

Additional artificial language results appear in Figures~\ref{fig:convergence-ranking}--\ref{fig:ppl_diff_5mb}; natural language results in Figures~\ref{fig:rightness-goldfish}, \ref{fig:goldfish_all_langs_directionality}, \ref{fig:combined-encoding-basewo}, \ref{fig:multilingual_combined}, and Table~\ref{tab:pud_bpec_by_order}.

\begin{table}[t]
\centering
\small
\begin{tabular}{lrcccc}
\toprule
\textbf{Order} & $n$ & \textbf{5\,MB} & \textbf{10\,MB} & \textbf{100\,MB} & \textbf{1\,GB} \\
\midrule
SVO         & 13 & 2.20 & 2.08 & \textbf{1.60} & \textbf{1.38} \\
SOV         &  4 & \textbf{2.16} & \textbf{2.05} & 1.67 & 1.49 \\
VSO         &  1 & 2.26 & 2.15 & 1.72 & 1.50 \\
NoDominant  &  2 & 2.21 & 2.08 & 1.61 & 1.38 \\
\bottomrule
\end{tabular}
\caption{Median BPEC of Goldfish models on PUD treebank languages by base word
order. SOV leads at small scales; SVO leads at large scales, mirroring
FLORES-200 (Table~\ref{tab:median-bpec-by-size}). Per-order samples are small
($n{=}1$--$13$), so we omit IQR and test the ordering directly: the SVO
advantage reaches significance by 1\,GB (SVO vs.\ SOV Mann--Whitney $p{=}.045$),
and the SOV$\times$size interaction replicates FLORES-200 (mixed-effects
$\beta{=}{+}0.04$, $p{=}.04$).}
\label{tab:pud_bpec_by_order}
\end{table}

\subsection{Artefact Use and Licensing}
All artefacts are used for research consistent with their 
intended purpose. FLORES-200 
\citep{nllbteam2022languageleftbehindscaling} and PUD 
\citep{zeman-etal-2017-conll}: CC-BY-SA~4.0; WALS 
\citep{dryer_order_2013} and Grambank 
\citep{skirgardGrambankRevealsImportance2023}: CC-BY~4.0; BLOOM 
\citep{workshop2023bloom176bparameteropenaccessmultilingual}: 
BigScience RAIL~v1.0; XGLM 
\citep{lin2022fewshotlearningmultilinguallanguage}: MIT; mGPT 
\citep{shliazhko-etal-2024-mgpt} and Stanza 
\citep{qi-etal-2020-stanza}: Apache~2.0; Wuggy 
\citep{keuleers-brysbaert-2010-wuggy}: GPL; wordfreq 
\citep{speer-2022-wordfreq}: MIT. Goldfish models
\citep{chang2026goldfishmonolinguallanguagemodels}
are publicly available but unlicensed.

\end{document}